# The Ultrametric Constraint and its Application to Phylogenetics

**Neil C.A. Moore**                                    NCAM@CS.ST-ANDREWS.AC.UK
*Computer Science, University of St. Andrews, Scotland*

**Patrick Prosser**                                    PAT@DCS.GLA.AC.UK
*Computing Science, Glasgow University, Scotland*

## Abstract

A phylogenetic tree shows the evolutionary relationships among species. Internal nodes of the tree represent speciation events and leaf nodes correspond to species. A goal of phylogenetics is to combine such trees into larger trees, called supertrees, whilst respecting the relationships in the original trees. A rooted tree exhibits an ultrametric property; that is, for any three leaves of the tree it must be that one pair has a deeper most recent common ancestor than the other pairs, or that all three have the same most recent common ancestor. This inspires a constraint programming encoding for rooted trees. We present an efficient constraint that enforces the ultrametric property over a symmetric array of constrained integer variables, with the inevitable property that the lower bounds of any three variables are mutually supportive. We show that this allows an efficient constraint-based solution to the supertree construction problem. We demonstrate that the versatility of constraint programming can be exploited to allow solutions to variants of the supertree construction problem.

## 1. Introduction

One of the grand challenges of phylogenetics is to build the *Tree of Life* (ToL), a representation of the evolutionary history of every living thing. To date, biologists have catalogued about 1.7 million species, yet estimates of the total number of species range from 4 to 100 million. Of the 1.7 million species identified only about 80,000 have been placed in the ToL so far (Pennisi, 2003). There are applications for the ToL: to help understand how pathogens become more virulent over time, how new diseases emerge, and to recognise species at risk of extinction (Pennisi, 2003; Mace, Gittleman, & Purvis, 2003). One approach to building the ToL is divide and conquer: combining smaller trees such as those available from TreeBase (TreeBASE, 2003) into so-called "supertrees" (Bininda-Emonds, 2004) to approach a more complete ToL.

To date, supertree construction has been dominated by imperative techniques (Semple & Steel, 2000; Semple, Daniel, Hordijk, Page, & Steel, 2004; Daniel, 2003; Bordewich, Evans, & Semple, 2006; Ng & Wormald, 1996; Bryant & Steel, 1995; Page, 2002) but recently new declarative approaches have emerged using constraint programming (Gent, Prosser, Smith, & Wei, 2003; Prosser, 2006; Beldiceanu, Flener, & Lorca, 2008) and answer set programming (Wu, You, & Lin, 2007). One of the properties of rooted trees that suits these approaches is that trees are by their nature *ultrametric*: in rooted trees the root node has depth 0, and the depth of other nodes is 1 plus the depth of their parent. Taking any three leaves $a$, $b$ and $c$ in pairs it must be that one pair has a deeper *most recent common ancestor* (mrca) than the other pairs, or that all three pairs have the same mrca. And this is what we mean by ultrametric, that for any three there is a tie for the minimum. In fact, if we know the depth of the mrca of all pairs of leaves the structure of the tree is uniquely determined. This inspires a constraint programming encoding for rooted trees, using the ultrametric constraint that we will define later. We explore solutions to the phylogenetic supertree problem and its variants. In so doing, we show the practicality of an ultrametric encoding for rooted tree problems, as well as arguing that this is a valuable addition to the set of techniques for supertree problems.

The paper is organised as follows. First, we introduce constraint programming and the supertree construction problem. We then propose a specialised ultrametric constraint, in terms of its propagation procedures, that maintains bounds(Z)-consistency (Bessière, 2006) on three variables. We show that this specialised constraint is required because models that use toolkit primitives cannot guarantee the ultrametric property on the supertree problem via propagation alone. Furthermore, the space complexity of such models becomes prohibitive. The ultrametric constraint is then extended to maintain the property over a symmetric matrix of variables. We then go on to show that this constraint can be efficiently applied to the problem of supertree construction, in particular that applying propagation to this model gives a polynomial time procedure for supertree construction. We then demonstrate this on real data and give justification that there has been an improvement in time and space over previous constraint encodings. One of the benefits of the constraint programming approach is that variants of the supertree problem can be addressed within this one model. We justify this assertion by proposing a constraint solution finding essential relations in the supertree (Daniel, 2003), addressing ancestral divergence dates (Semple et al., 2004; Bryant, Semple, & Steel, 2004), modelling nested taxa (Page, 2004; Daniel & Semple, 2004) and coping with conflicting data.

## 2. Background

In this section we give necessary definitions and descriptions of the Constraint Satisfaction Problem (Tsang, 1993), Constraint Programming, and the Supertree problem.

### 2.1 Constraint Programming and the CSP

Constraint Programming (CP) (Rossi, van Beek, & Walsh, 2007) is a declarative style of programming where problems are modelled as a CSP, i.e., as a set of variables that have to be assigned values from those variables' domains to satisfy a set of constraints. Values might typically be integers drawn from finite domains, real numbers from ranges, or more complex entities like sets or graphs. We will only be considering integers.

**Definition 1.** *A constraint satisfaction problem (CSP) is a triple $(V, D, C)$ where $V$ is a set of $n$ variables $\{v_1, \ldots, v_n\}$; $D = \{dom(v_1), \ldots, dom(v_n)\}$ is a collection of domains, each a totally ordered set of integer values; and $C = \{c_1, \ldots, c_e\}$ is a set of $e$ constraints, each with a scope of variables $scope(c) = (v_{c_1}, \ldots, v_{c_k})$ and a relation $rel(c) \subseteq dom(v_{c_1}) \times \ldots \times dom(v_{c_k})$. An assignment of value $x \in dom(v)$ to variable $v_i \in V$ is denoted by $(v_i, x)$. A constraint $c \in C$ is satisfied by an assignment $\{(v_{c_1}, x_{c_1}), \ldots, (v_{c_k}, x_{c_k})\}$ when $scope(c) = (v_{c_1}, \ldots, v_{c_k})$* and $(x_{c_1}, \ldots, x_{c_k}) \in rel(c)$. *A set of assignments* $\{(v_1, x_1), \ldots, (v_n, x_n)\}$ *involving every variable in the problem is a* solution *when it satisfies all the constraints in* $C$.

A constraint solver finds a solution to a CSP via a process of *constraint propagation* and search. Constraint propagation is an inferencing process that takes place when a variable is initialised or loses values. Propagation maintains a level of consistency, such as arc-consistency (Mackworth, 1977), across the variables, removing values from domains that cannot occur in any solution (i.e., removing unsupported values). We use the definitions of (generalized) arc-consistency ((G)AC) due to Bessière (2006):

**Definition 2.** *Given a CSP* $(V, D, C)$, *a constraint* $c \in C$ *with* $scope(c) = (v_{c_1}, \ldots, v_{c_k})$, *and a variable* $v \in scope(c)$, *a value* $x \in dom(v)$ *is consistent with respect to* $c$ *(alternatively, supported by* $c$*) iff there exists a satisfying assignment* $\alpha = \{(v_{c_1}, a_1), \ldots, (v_{c_k}, a_k)\}$ *for* $c$ *such that* $(v, x) \in \alpha$ *and* $\forall i,\ a_i \in dom(v_{c_i})$. *The domain* $dom(v)$ *is (generalized) arc-consistent on* $c$ *iff all values in* $dom(v)$ *are consistent with respect to* $c$, *and the CSP is (generalized) arc-consistent when all variable domains are (generalized) arc-consistent on all constraints in* $C$.

Arc-consistency can be established on a CSP using an algorithm such as AC3 (Mackworth, 1977). For sake of exposition we assume all constraints in $C$ are binary and that for each constraint $c$ we have a counterpart $c'$ such that $scope(c) = (v_a, v_b)$ and $scope(c') = (v_b, v_a)$ with $rel(c') = rel^{-1}(c)$. For example, if we have the constraint $c_{xy} = x < y$ then we also have the constraint $c_{yx} = y > x$. At the heart of AC3 is the *revise* function, which takes a binary constraint $c$ as its argument and delivers a Boolean result. The function removes from $dom(v_a)$ all values that have no support in $dom(v_b)$ w.r.t. the constraint $c$, and returns true if any removals take place. Initially all constraints are added to a set $S$. Constraints are iteratively removed from $S$ and revised. If $revise(c_{km})$ returns true then $S$ becomes $S \cup \{c_{ik} | c_{ik} \in C \land i \neq k \land i \neq m\}$. This step can be considered as the propagation of a domain reduction on variable $v_k$ to variables constrained by $v_k$. The iteration terminates when $S$ is empty or a variable's domain becomes empty. When $S$ is empty the arc-consistency algorithm has reached a fixed point (i.e., a further application of the arc-consistency process will have no effect on the domains of the variables) and the problem has been made arc-consistent. When a domain empties, we have shown the there are no solutions globally and hence we can stop. The AC3 algorithm has $O(e \cdot d^3)$ time complexity, where $e$ is the number of constraints and $d$ the size of the largest domain, however other algorithms can achieve a time bound of $O(e \cdot d^2)$ (Yuanlin & Yap, 2001; Bessière & Régin, 2001).

We demonstrate arc-consistency with the example of Figure 1 by Smith (1995). We have three constrained integer variables $x$, $y$ and $z$, each with an integer domain $\{1..5\}$, and binary constraints $c_{xy} : x < y - 2$, $c_{yz} : y + z$ is even, and $c_{zx} : z < 2x + 1$. Since the constraints are binary we can represent the problem as a constraint graph, where nodes are vertices and edges are constraints. Initially the constraint $c_{xy}$ is revised with respect to $x$ and the values $\{3..5\}$ are removed from $dom(x)$. Then $c_{xy}$ is revised w.r.t. $y$ and $dom(y)$ becomes $\{4, 5\}$. $c_{yz}$ is then revised w.r.t. $y$ with no effect and then revised w.r.t. $z$, again with no effect. Revising $c_{zx}$ w.r.t. $z$ reduces $dom(z)$ such that it becomes $\{1..4\}$, consequently the constraint $c_{yz}$ is added into the set of constraints pending revision. Constraint $c_{zx}$ is

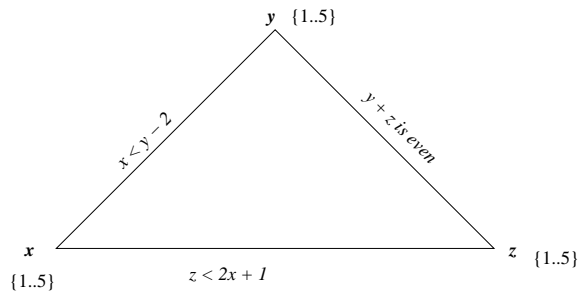

Figure 1: A binary constraint satisfaction problem.

```
import choco.Problem;
import choco.ContradictionException;
import choco.integer.*;

public class BMStut {

    public static void main(String[] args)  throws ContradictionException {
        Problem pb      = new Problem();
        IntDomainVar x    = pb.makeEnumIntVar("x",1,5); // x in {1..5}
        IntDomainVar y    = pb.makeEnumIntVar("y",1,5); // y in {1..5}
        IntDomainVar z    = pb.makeEnumIntVar("z",1,5); // y in {1..5}
        IntDomainVar even = pb.makeEnumIntVar("even",new int[] {2,4,6,8,10});

        pb.post(pb.gt(pb.minus(y,2),x));          // x < y - 2
        pb.post(pb.gt(pb.plus(pb.mult(2,x),1),z)); // z < 2x + 1
        pb.post(pb.eq(even,pb.plus(y,z)));        // y + z is even

        pb.solve();                               // solve using MAC
    }
}
```

Figure 2: A JChoco constraint program for the CSP of Figure 1.

revised w.r.t. $x$ and then $c_{yz}$ w.r.t. to $y$, both with no effect. The revision set at that point is empty and arc-consistency has been established with variable domains $dom(x) = \{1, 2\}$, $dom(y) = \{4, 5\}$, and $dom(z) = \{1..4\}$.

Solving a CSP may involve search, i.e., we might need to try different values for variables in order to determine if a solution exists. Typically a constraint solver will begin by establishing arc-consistency, and then repeatedly select a variable and assign it a value from its domain (*instantiate* it). This effectively reduces that variable's domain to a singleton, and arc-consistency is then re-established. If this succeeds another instantiation is made, but if it fails we backtrack by undoing the most recent instantiation. This is called MAC, for maintaining arc-consistency (Sabin & Freuder, 1994).

Figure 2 shows a constraint program for the problem in Figure 1 using the choco constraint programming toolkit for the Java language (Choco, 2008), and it finds solution $x = 1$, $y = 4$, and $z = 2$ first.

Constraint toolkits tend to be based around the AC5 algorithm (van Hentenryck, Deville, & Teng, 1992), allowing propagators to be specialised for specific constraints resulting in improved efficiency and adaptability. AC5 amends set $S$ from AC3 to contain triples of the form $(v, c, \delta)$ where $v \in scope(c)$ and $\delta$ is the set of values lost by $v$, consequently revision is more efficient because propagation can focus of values that may have lost support, rather than having to check every value for support. In an object-oriented toolkit language a constraint has associated propagation methods that should be implemented, and these methods are activated when a domain event occurs on a variable involved in that constraint. Domain events can be the initialisation of a variable, an increase in the lower bound, a decrease in the upper bound, the removal of a value between the bounds, or the instantiation of that variable. This is an exhaustive list, however some toolkits allow only one event: that one or more values have been lost and the propagator writer must then determine what action to take. To give examples of using a toolkit with specialised constraints, when modelling a routing problem we might have a constrained integer variable for each location to be visited, with a domain of values corresponding to the index of the next destination (the so-called "single-successor" model). A subtour elimination constraint (Caseau & Laburthe, 1997) might then be used to ensure that only legal tours are produced, and Régin's all-different constraint (Régin, 1994) could be added to increase domain filtering. In a pick up and delivery variant, side constraints could be added to ensure that some locations are visited before others. For a job shop scheduling problem we might have a model that uses 0/1 variables to decide the relative order of pairs of activities that share a resource, and we might increase propagation by adding Carlier and Pinson's edge finding constraint (1994).

The constraint programming approach is general and practical for modelling and solving problems, and provides a framework for the combination of problem specific algorithms in one solver. This allows us to solve many classes of problems efficiently and to model even more problems via the addition of side constraints.

## 2.2 The Supertree Problem

Supertree construction is a problem in phylogenetics where we are to combine leaf-labelled species trees, where the sets of leaf labels intersect, into a single tree that respects all arboreal relationships in each input tree (Bininda-Emonds, 2004). Species trees describe part of the evolutionary history of a set of species. Labels on leaves correspond to existing species and internal nodes represent divergence events in evolutionary history where one species split into at least two other species. Species trees may also be annotated with dates on internal nodes, representing the time at which the divergence event happened.

We now define the term *displays*, which makes precise what we mean by "respects arboreal relationships": supertree $T_1$ *displays* a tree $T_2$ if and only if $T_2$ is equivalent to $T_4$ (i.e. they induce the same hierarchy on the leaf labels) where $T_4$ is obtained by the following steps (Semple & Steel, 2000):

1. Let $L$ be the set of leaves of $T_1$ that are in $T_2$.
2. Let $T_3$ be the unique subtree of $T_1$ that connects all leaves in $L$.
3. To obtain $T_4$: wherever there is a subpath $(p_1, \ldots, p_k)$ of a path from the root to a leaf in $T_3$ where $p_2, \ldots, p_{k-1}$ are all interior nodes of degree 2, contract this into a single edge.

The problem is then to produce a rooted species tree $T$ from a forest of input trees F, so that $T$ contains all the species in $F$ and displays every tree in $F$. Figures 3 and 4 illustrate the displays property.

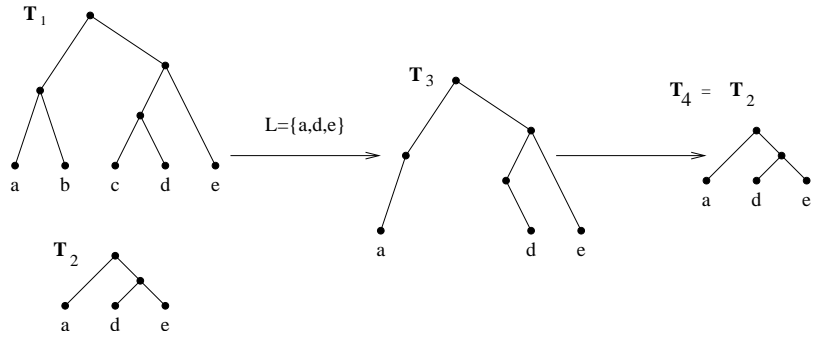

Figure 3: An example of a tree $T_1$ that displays a tree $T_2$

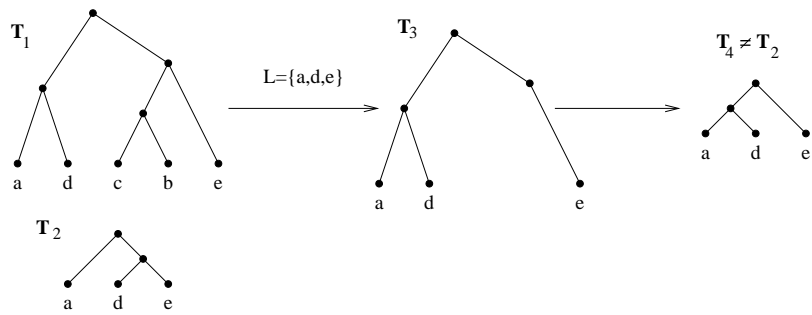

Figure 4: An example of a tree $T_1$ that does not display tree $T_2$

We say that two trees $T_1$ and $T_2$ are compatible (incompatible) if there exists (doesn't exist) a third tree $T_3$ that displays $T_1$ and $T_2$. Variants on the supertree problem that have previously been published and solved in the specialist bioinformatics literature include finding all solutions[1], counting solutions, finding conserved relationships in all supertrees (Daniel, 2003), incorporating nested taxa (Semple et al., 2004), incorporating ancestral divergence dates (Semple et al., 2004) and the possibility of contradictory input data (Semple & Steel, 2000).

## 3. The Ultrametric Constraint

The ultrametric constraint was first proposed by Gent et al. (2003) within the context of supertree construction (Bininda-Emonds, 2004), and was implemented using toolkit primitives. We review this encoding and show that in most constraint toolkits this is inefficient in terms of both space and time. This motivates the creation of a specialised ultrametric propagator over three variables, that maintains the ultrametric property over the bounds of those variables. It is presented by describing the necessary propagation methods. We then extend it to a specialised propagator that maintains the ultrametric property on a symmetric matrix of variables.

### 3.1 Previous Work on the Ultrametric Constraint

First, we give a definition of the ultrametric constraint.

**Definition 3.** *An* ultrametric constraint *on three variables (henceforth, Um-3) x, y and z constrains them such that:*

$$(x > y = z) \ \lor \ (y > x = z) \ \lor \ (z > x = y) \ \lor \ (x = y = z) \tag{1}$$

This constraint ensures that there is a tie for the least element of the three, i.e., either all three are the same, or two are the same and the other is greater. The constraint was proposed by Gent et al. (2003), used again by Prosser (2006) and both times implemented as a literal translation of Equation 1 using toolkit primitives. Evidence obtained from the JChoco, ECLiPSe and ILog constraint programming toolkits shows that no propagation is done to lower bounds by this combination of primitive constraints. This is due to the disjunctive constraints since in many constraint programming toolkits propagation is delayed until only one of the disjuncts can be true, this is known as *delayed-disjunction* consistency (van Hentenryck, Saraswat, & Deville, 1998). Consequently, in the above encoding values that cannot occur in any satisfying assignment might not be pruned from the domain of a variable. Consider the case for three variables: $x \in \{1, 2, 3\}$, $y \in \{2, 3\}$ and $z \in \{3\}$. The domains of the variables are already at a fixed point with respect to delayed-disjunction consistency but there is no ultrametric assignment where $x$ takes the value 1, i.e., delayed-disjunction propagation does not achieve arc-consistency. As we shall see later, finding a solution to the supertree problem using toolkit constraints can result in a backtracking search, and we prefer to avoid this. Of course, higher levels of consistency would overcome this, such as constructive-disjunction consistency (van Hentenryck et al., 1998), singleton

---

1. There may be multiple supertrees for the same set of input trees.

arc-consistency (Debruyne & Bessière, 1997) or the filtering algorithm of Lhomme (2003). However, the cost of these is greater in the average case than delayed-disjunction, preventing their use in toolkits. In fact for the UM-3 constraint it is especially unfortunate that the lower bounds may not be trimmed properly:

**Lemma 1.** *In the* UM-3 *constraint, when lower bounds are supported (i.e., form an ultra-metric instantiation with values in the other constrained variables), they support each other.*

*Proof.* Consider three supported lower bounds. Suppose for a contradiction that the two least of these are distinct. One of these is distinct lowest and it cannot be supported on account of the fact that it is not equal to anything or larger than anything. Therefore by contradiction the two least must be equal. However the other lower bound is at least as large as these, so the lower bounds are mutually supportive. ▢

This Lemma will have important implications for the species tree model to be presented in detail in Section 4: in particular, the lower bounds of a bounds(Z)-consistency model form a solution, where bounds(Z)-consistency (Bessière, 2006) is defined as follows

**Definition 4.** *Given* $(V, D, C)$ *and constraint* $c \in C$ *with* $scope(c) = (v_{c_1}, \ldots, v_{c_k})$, *a tuple* $\tau = (x_{c_1}, \ldots, x_{c_k})$ *is a bound support when* $\tau \in rel(c)$ *and for all* $x_{c_i} \in \tau$, $\min(dom(v_{c_i})) \leq x_{c_i} \leq \max(dom(v_{c_i}))$. *A constraint* $c$ *is bounds(Z)-consistent if for all* $v_{c_i} \in scope(c)$ *there exist bound supports involving both* $\min(dom(v_{c_i}))$ *and* $\max(dom(v_{c_i}))$. *A CSP is bounds(Z)-consistent when every constraint* $c \in C$ *is bounds(Z)-consistent.*

We will henceforth abbreviate "bound support" to "support" and bounds(Z)-consistency to BC(Z). BC(Z) differs from AC because it puts weaker conditions on the values that comprise the support: rather than them having to be in the domain, they need to only be between the lower and upper bounds of the domain. This means that BC(Z) prunes a subset of the values that AC can, in general. Weaker levels of consistency such as BC(Z) are useful because in certain problems they can prune the same number of values as AC more easily, or fewer values much more quickly. In our case, BC(Z) is interesting because this level of consistency is "just enough" to ensure that the problem can be solved by propagation with no search, as we shall see.

## 3.2 Design of a BC(Z) UM-3 Propagator

In this Section we describe a UM-3 propagator that enforces BC(Z), namely UM-3-BCZ.

### 3.2.1 Analysis of Lower and Upper Bounds

In this section we don't take account of domains becoming empty. Since we analyse lower and upper bounds in isolation a lower bound may pass an upper bound or vice-versa, thereby emptying a domain. If this happens then the propagator described in Section 3.2.2 will not enforce BC(Z), rather it will terminate. This is not a problem, because a domain becoming empty means that there is no solution and to continue would be a waste of time. Concordantly, in this section, when we analyse lower bounds we will assume the upper bound is $\infty$ so that the domain cannot become null for this reason, and vice-versa.

Code style: Variables $v_1$, $v_2$, $v_3$, $S$, $M$ and $L$ below are constrained integer variables, and are synonymous with their domains. Consequently a variable $x$ can be considered as a domain where $x$.**lb** and $x$.**ub** return references to lower and upper bounds respectively; **SortOnLowerBounds**($x$,$y$,$z$) returns a tuple of references to variables $x$, $y$ and $z$ in non-decreasing order of their lower bounds; **SortOnUpperBounds** is analogous; **let** $(S,M,L) \leftarrow \ldots$ names the references $S$, $M$ and $L$; as a result of $S$.**lb** $\leftarrow M$.**lb**, the lower bound of $S$ is assigned equal to the value of the lower bound of $M$, although if $M$.lb subsequently changes they will be distinct again; the expression $x \cap_b y$ returns the intersection of their domains [max(x.lb,y.lb) ... min(x.up,y.up)].

Algorithm UM-3-BCZ

| | |
|---|---|
| **LBFix**($v_1$,$v_2$,$v_3$) | |
| A1 | **let** $(S, M, L) \leftarrow$ SortOnLowerBounds($v_1, v_2, v_3$) |
| A2 | **if** ($S$.lb $< M$.lb) **then** |
| A2.1 | $S$.lb $\leftarrow M$.lb |
| **UBFix**($v_1$,$v_2$,$v_3$) | |
| A3 | **let** $(S, M, L) \leftarrow$ SortOnUpperBounds($v_1, v_2, v_3$) |
| A4 | **if** ($S$.ub $< M$.ub) **then** |
| A4.1 | **if** ($S \cap_b L = \emptyset$) **then** |
| A4.2 | $M$.ub $\leftarrow S$.ub |
| A4.3 | **else if** ($S \cap_b M = \emptyset$) **then** |
| A4.4 | $L$.ub $\leftarrow S$.ub |
| **Min event**($v_1$,$v_2$,$v_3$) | |
| A5 | LBFix($v_1$,$v_2$,$v_3$) |
| A6 | **if** all domains are non-empty **then** |
| A6.1 | UBFix($v_1$,$v_2$,$v_3$) |
| **Max event**($v_1$,$v_2$,$v_3$) | |
| A7 | UBFix($v_1$,$v_2$,$v_3$) |
| **Fix event**($v_1$,$v_2$,$v_3$) | |
| A8 | LBFix($v_1$,$v_2$,$v_3$) |
| A9 | **if** all domains are non-empty **then** |
| A9.1 | UBFix($v_1$,$v_2$,$v_3$) |

Figure 5: Algorithm for UM-3-BCZ propagator

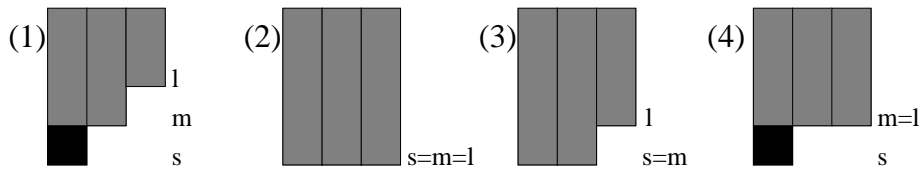

Figure 6: Cases in the analysis of LBFix

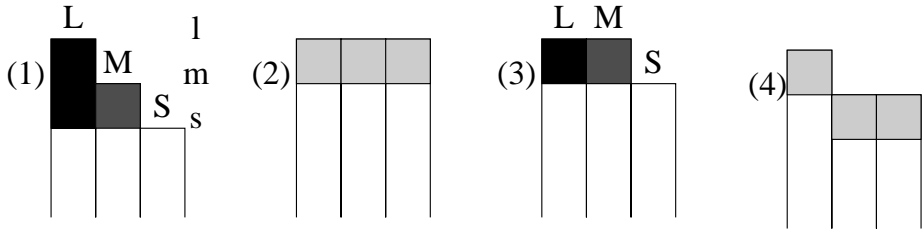

Figure 7: Cases in the analysis of UBFix

The procedure LBFix in Figure 5 takes as input three variables and removes any unsupported values at the lower bounds of the domain. The intuition for the algorithm achieving this is that each one needs to be involved in a tie for least element, hence if the smallest lower bound is strictly less than the others then it must be unsupported.

The possible states of lower bounds when LBFix is invoked are summarised in Figure 6. Either all three are different (case 1), all three are the same (case 2) or two are the same and one different (cases 3 and 4). These give relationships between bounds at a point in time when some lower bound may be unsupported. The boxes in Figure 6 are shaded as follows: regions shaded black are removed by propagation whereas gray regions are supported. What the diagrams are *not* supposed to suggest is that, for example in case 1, the bounds differ by 1. Rather when two bounds are lined up they are the same bound and when one is different from another they are different by some non-zero but unspecified amount. Hence they describe relationships and not actual values.

The following shows that LBFix removes all unsupported values and does not remove any supported values.

**Lemma 2.** *After LBFix is invoked all lower bounds of the argument variables are supported w.r.t. the* UM-3 *constraint and no supported values are removed.*

*Proof.* For cases 1 and 4 (Figure 6) the condition on line A2 is satisfied, so line A2.1 is executed, this results in the removal of the unsupported range and by inspection the remaining bounds are mutually supportive. In cases 2 and 3, the condition on line A2 is failed and so no changes are made to the domains; the bounds are mutually supportive. □

The procedure UBFix in Figure 5 does the same job to upper bounds that LBFix does to lower bounds. The following Lemma justifies this assertion and the cases used in the proof are shown in Figure 7:

**Lemma 3.** *After UBFix is invoked either*

- *all upper bounds of the argument variables are supported w.r.t. the UM-3-BCZ constraint and no supported values are removed, or*

- *a domain is null as a result of removing unsupported values.*

*Proof.* Let $S$, $M$ and $L$ be the domains with smallest, middle and largest upper bounds, breaking ties arbitrarily. Let $s$, $m$ and $l$ be these upper bounds.

First we will prove that in case 1 (Figure 7) the shaded region in $M$ is supported if and only if $L \cap_b S \neq \emptyset$: Potentially, the bound can be supported by

- equal values in $S$ and $L$ at least as small as it (i.e., $S \cap_b L \neq \emptyset$), or

- an equal value in either $S$ or $L$, and a value at least as large in the remaining domain.

However, notice that the latter is impossible due to the fact that only $L$ contains an equal value, and $S$ has no value as large as this.

Similar arguments can establish that the shaded regions in $L$ in case 1, $M$ in case 3 and $L$ in case 3 are supported if and only if $M \cap_b S \neq \emptyset$, $L \cap_b S \neq \emptyset$ and $M \cap_b S \neq \emptyset$, respectively.

Now we will establish the Lemma for each of the cases 1, 2, 3 and 4:

**Cases 2 and 4** The condition on line A4 is false, and so no domains are changed. The upper bounds are mutually supportive in each case.

**Case 1 and shaded region of $M$ is unsupported** From above $L \cap_b S = \emptyset$. Hence UBFix line A4.2 will be executed and the unsupported region removed. Now the upper bounds $l \in L$, $s \in M$ and $s \in S$ are mutually supportive.

**Case 1 and shaded region of $M$ is supported** From above $L \cap_b S \neq \emptyset$. If the shaded region of $L$ is also supported then $M \cap_b S \neq \emptyset$ and so neither line A4.2 nor A4.4 is executed and no changes are made to the domains. The upper bound of $S$ is also supported, by $m \in L$, $m \in M$ and $s \in S$. If the shaded region of $L$ is *not* supported then $M \cap_b S = \emptyset$ and so line A4.4 is executed resulting in the removal of the region. The new bounds $s \in S$, $m \in M$ and $s \in L$ are mutually supportive.

**Case 3 and shaded regions of $M$ and $L$ supported** From above, $L \cap_b S \neq \emptyset$ and $M \cap_b S \neq \emptyset$. Hence no domain changes result from executing UBFix. $l \in L$ and $s \in S$ are supported by $l \in L$, $s \in M$ and $s \in S$. $m \in M$ is supported by $s \in L$, $m \in M$ and $s \in S$.

**Case 3 and shaded regions of $M$ and $L$ are unsupported** From above, $L \cap_b S = \emptyset$ and $M \cap_b S = \emptyset$ and so line A4.2 of UBFix is executed and this results in $M$ becoming null.

**Case 3, shaded region of $M$ supported but shaded region of $L$ not supported** From above $L \cap_b S \neq \emptyset$ and $M \cap_b S = \emptyset$, so that A4.4 is executed to remove the unsupported region. The new bounds of $s \in S$, $m \in M$ and $s \in L$ are mutually supportive.

**Case 3, shaded region of $L$ supported but shaded region of $M$ not supported** Symmetric with previous case.                                                                                    □

Note that there is no analog of Lemma 1 for upper bounds since, for example, the bounds of $x = \{1, 2, 3\}$, $y = \{1, 2\}$ and $z = \{1\}$ are all supported, but *not* mutually supportive.

### 3.2.2 THE PROPAGATION ALGORITHM

Having presented LBFix and UBFix we are now in a position to present the complete propagation algorithm. The propagator works with arbitrary domains and it enforces BC(Z), except when a domain becomes empty, in which case it does no further work. The algorithm is described by the action taken when any of three domain events occur:

**min** The domain has lost its lower bound since propagator was last invoked.

**max** The domain has lost its upper bound since propagator was last invoked.

**fix** The domain is a singleton, i.e., the variable is instantiated and upper and lower bounds are equal.

i.e. we only consider events on the bounds of the variables. The algorithm is listed in lines A5-A9 of Figure 5. Intuitively these procedures work because, as we will show, a change to an upper bound can affect the support for other upper bounds, but a change in a lower bound can affect support for both lower and other upper bounds. Hence we need only run LBFix when a lower bound may have changed, but UBFix must be run for a change of either lower or upper bounds. Whilst it would be correct to cycle between trimming upper and lower bounds until a fixed point is reached (i.e. no more changes occur), we can guarantee a fixed point more easily.

**Lemma 4.** *It is possible for a change in a lower bound to result in the loss of support for another lower bound.*

*Proof.* All the bounds in the diagram have support, but when the black shaded lower bound is lost the dark gray shaded lower bound loses support.

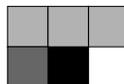

□

**Lemma 5.** *It is possible for a change in a lower bound to result in the loss of support for an upper bound.*

*Proof.* All the bounds in the diagram have support, but when the black shaded lower bound is lost the dark gray shaded upper bound loses support.

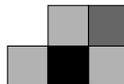

□

**Lemma 6.** *It is possible for the loss of an upper bound to cause the loss of support for another upper bound.*

*Proof.* All the bounds in the diagram have support, but when the black shaded upper bound is lost the dark gray shaded upper bound loses support.

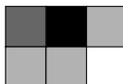

□

**Corollary 1.** *It is impossible for a change in an upper bound to result in the loss of support for a lower bound.*

*Proof.* By Lemma 1 a lower bound retains support as long as the other lower bounds are intact, hence losing an upper bound has no effect. □

Why the asymmetry between upper and lower bounds? It is due to the asymmetry in the definition of UM-3-BCZ and has the practical repercussion that BC(Z) lower bounds must be mutually supportive whereas BC(Z) upper bounds may not be and may require support from other values including lower bounds.

Corollary 1 suggests that a further improvement on the algorithm in Figure 5 is to execute Line A6 and A9 if and only if any lower bound lost is the only remaining support for an upper bound. However, the conditionals intrinsic in UBFix amount to much the same thing and there is little point in repeating them.

The point of these theorems has been to build up a complete proof of correctness and BC(Z) status:

**Theorem 1.** *The code for min, max and fix events listed in Figure 5 does not remove any values involved in bound supports for the UM-3-BCZ constraint, and, if all the domains are non-null after propagation, the resultant domains will be BC(Z).*

*Proof.* First we must establish that no values are removed during propagation that could be involved in a support, and that the result domains are subsets of the input domains. The former is immediate from Lemmas 2 and 3, because values are only removed as a result of executing LBFix and UBFix. The latter is immediate from inspection of LBFix and UBFix, because they only ever make lower bounds larger and upper bounds smaller.

The final thing to establish is that BC(Z) is enforced, unless a domain becomes null.

If any domain becomes empty as a result of running the algorithm then the Theorem is trivially true.

If no domain becomes empty then we must show that all the bounds are supported. For lower bounds, by Lemma 4 and Corollary 1 we know that only the loss of a lower bound can result in the need to change a lower bound during propagation. Lower bounds can change as a result of either fix or min events, hence the propagator in Figure 5 runs LBFix in either event. When LBFix runs it leaves all lower bounds supported, as was shown in Lemma 2. For upper bounds, by Lemmas 5 and 6 we know that the loss of either a lower or upper bound can result in the loss of an upper bound. Hence upper or lower bounds can change as a result of any event, and lower bounds can also change as a result of LBFix, hence the propagator runs UBFix in all events, and it runs after LBFix has finished, if necessary. When UBFix runs it leaves all upper bounds supported provided no domain becomes empty, as was shown in Lemma 3. □

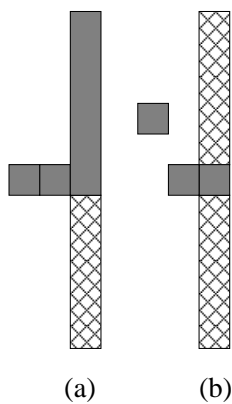

Figure 8: Propagation done when 2 domains are singleton

The propagation algorithm runs in $\Theta(1)$ time. This is because all of the operations in LBFix, UBFix, min, max and fix events are $\Theta(1)$, provided that our domain representation allows access to the upper and lower bounds in $\Theta(1)$. This can be guaranteed if domain reductions only occur at the bounds, as is the case here, or if domains are represented using one of the structures proposed by van Hentenryck et al. (1992).

## 3.3 Entailment

Schulte and Carlsson (2006) define entailment as when all possible constractions of the domains in a constraint's scope are consistent. If we can detect that this has happened we can stop running the propagator henceforth, since it cannot prune any more values.

**Definition 5.** *A propagator is* entailed *by domains $D = \{d_1, \ldots, d_n\}$ when any set of domains that are subsets of these, i.e., any $E = \{e_1, \ldots, e_n\}$ s.t. $\forall i.e_i \subseteq d_i$, are at a fixed point.*

We now describe a sufficient condition for the UM-3-BCZ constraint to be entailed, i.e., the UM-3-BCZ constraint becomes entailed as soon as two variables have singleton domains:

**Theorem 2.** UM-3-BCZ *becomes entailed as soon as two variables have singleton domains.*

*Proof.* Consider the possible scenarios: either the two singletons are the same (case (a) in Figure 8) or they are distinct (case (b) in Figure 8). The domains before propagation are shown in Figure 8 as boxes; the domains after propagation are shaded gray. Clearly all remaining choices for the third variable are valid instantiations and since the propagation algorithm is safe they cannot be removed by further propagation and by definition the propagation will be at a fixed point. □

## 3.4 Ultrametric Matrix Constraint

The supertree model presented in Section 4 makes use of the ultrametric constraint, however in this context the desired end product is to constrain a whole matrix to be an *ultrametric matrix*, and not merely to constrain three variables.

Code style: **let** $(i, j) \leftarrow$ index$(v)$ declares $i$ and $j$ to be the indices of variable $v$ in the matrix $M$ that the constraint is over.

| Algorithm UM-Matrix-BCZ |
| --- |
| **Min event**$(v)$ |
| A1     **let** $(i, j) \leftarrow$ index$(v)$ |
| A2     **for** $k \leftarrow 1 \ldots n$ **do** |
| A2.1       **if** $k \neq i$ and $k \neq j$ **then** |
| A2.2         Min event$(M_{ij}, M_{ik}, M_{jk})$ |
| A2.3         Max event$(M_{ij}, M_{ik}, M_{jk})$ |
| **Max event**$(v)$ |
| A3     **let** $(i, j) \leftarrow$ index$(v)$ |
| A4     **for** $k \leftarrow 1 \ldots n$ **do** |
| A4.1       **if** $k \neq i$ and $k \neq j$ **then** |
| A4.2         Max event$(M_{ij}, M_{ik}, M_{jk})$ |
| **Fix event**$(v)$ |
| A5     **let** $(i, j) \leftarrow$ index$(v)$ |
| A6     **for** $k \leftarrow 1 \ldots n$ **do** |
| A6.1       **if** $k \neq i$ and $k \neq j$ **then** |
| A6.2         Min event$(M_{ij}, M_{ik}, M_{jk})$ |
| A6.3         Max event$(M_{ij}, M_{ik}, M_{jk})$ |

Figure 9: Algorithm for UM-Matrix-BCZ propagator

**Definition 6.** *A symmetric matrix $M$ is an ultrametric matrix if and only if for every set of three distinct indices $i$, $j$ and $k$, there is a tie for the minimum of $M_{ij}$, $M_{ik}$ and $M_{jk}$; and $M_{ii} = 0$ for all $i$.*

The ultrametric matrix constraint can be achieved for matrix $M$ by posting the constraint UM-3-BCZ$(M_{ij}, M_{ik}, M_{jk})$ over all choices of distinct $i$, $j$ and $k$, but at a cost of introducing $\binom{n}{3}$ constraints. In practical constraint solvers any model containing this constraint will have $\Omega(n^3)$ space complexity, since the solver must have a list of all $\Theta(n^3)$ constraints stored somewhere. However, when a domain event occurs for any matrix variable $M_{ij}$ it is straightforward to iterate over all $k$ indices doing the same propagation as UM-Matrix-BCZ from Figure 5. This replaces a $\Theta(n^3)$ space list representation of the set of UM-3-BCZ constraints by a $\Theta(1)$ code representation. Hence we propose the ultrametric matrix constraint propagator UM-Matrix-BCZ in Figure 9.

This propagator mimics part of the AC3 algorithm (Mackworth, 1977) since it (a) receives a propagation event on a variable, (b) identifies which constraints are over that variable, and (c) arranges for the propagation to be carried out. Any events caused as a result are queued and dispatched by the underlying propagator as normal and this may cause UM-Matrix-BCZ to be run again. Each variable can be involved in up to $n - 2$ constraints, since each variable has *two* indices in the matrix and we have a constraint involving each choice of three different indices.

The algorithm propagates in $\Theta(n)$ time, which is more expensive per event than using $\binom{n}{3}$ Um-3 constraints, but a factor of $n$ fewer propagators wake up as a result of each event.

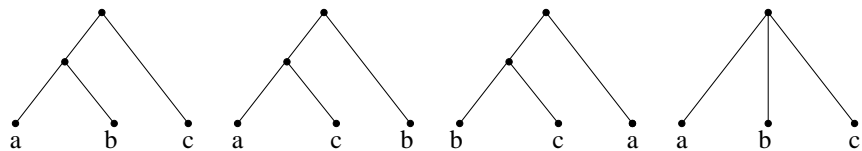

Figure 10: The four possible relationships between three leaf nodes in a tree: i.e. the three triples $(ab)c$, $(ac)b$, and $(bc)a$, and the fan $(abc)$.

## 4. Supertree Construction

We now review imperative solutions to the supertree construction problem, review the first constraint programing solution (Gent et al., 2003), and present a new encoding that exploits the specialised UM-MATRIX-BCZ constraint.

### 4.1 Imperative Solutions to the Supertree Problem

The earliest imperative techniques are due to Bryant and Steel (1995) and Ng and Wormald (1996). Both present a ONETREE algorithm which is based on the BUILD algorithm of Aho, Sagiv, Szymanski, and Ullman (1981). ONETREE is based on the observation that in a tree any three leaf nodes define a unique relation with respect to their most recent common ancestor (mrca), such that $mrca(a, b)$ is the interior node furthest from the root that has both leaf nodes $a$ and $b$ as descendants. We abuse notation by writing $mrca(a, b) > mrca(c, d)$ when the former has a greater depth than the latter, and similarly $mrca(a, b) = mrca(c, d)$ if they have the same depth. Given three different leaf nodes/species (labelled a, b, and c) one of the following four relations must hold:

$$(1) \quad mrca(a, b) > mrca(a, c) = mrca(b, c)$$
$$(2) \quad mrca(a, c) > mrca(a, b) = mrca(c, b)$$
$$(3) \quad mrca(b, c) > mrca(b, a) = mrca(c, a)$$
$$(4) \quad mrca(a, b) = mrca(a, c) = mrca(b, c)$$

We now say that in (1), (2) and (3) we have the *triples* $(ab)c$, $(ac)b$, and $(bc)a$ (where $(xy)z$ can be read as "x is closer to y than z") and in (4) we have the *fan* $(abc)$, i.e., in a fan the relationship between species is unresolved as we dont specify which pair is most closely related. This is shown in Figure 10. Prior to applying the ONETREE algorithm two (or more) species trees are broken up into triples and fans using the BREAKUP algorithm (Ng & Wormald, 1996), resulting in a linear sized encoding of those trees. The supertree is then constructed (if possible) using this encoding as input.

Figure 11 shows an example of the BREAKUP algorithm process. Two variants of the process are shown; at the top we have a hard breakup, where fans are considered as hard evidence that must be respected (*hard polytomies* as described by Ng and Wormald, 1996) and below a soft breakup where fans are taken as a lack of evidence (*soft polytomies* as described by Bryant and Steel, 1995). For hard breakup the algorithm is modified such

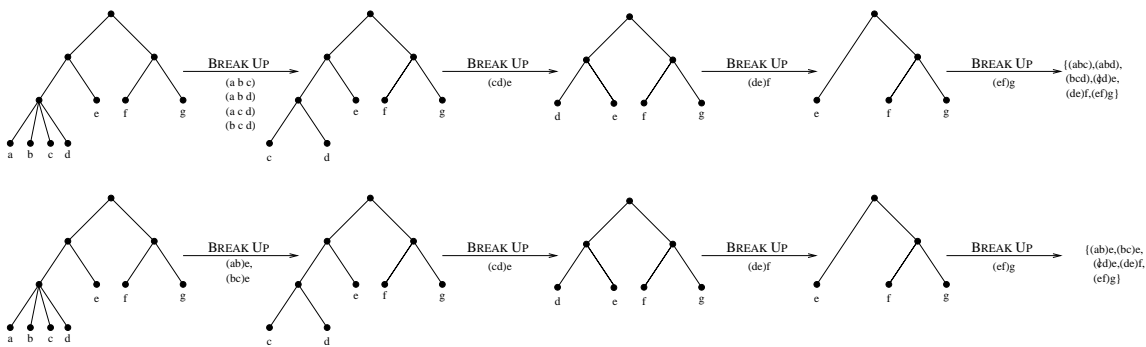

Figure 11: Example execution of the BreakUp algorithm. On the top, a hard breakup, and below a soft breakup (no fans produced)

Code style: The function sortedInteriorNodes($T$) delivers the set of interior nodes of the tree $T$ in non-increasing order of depth in that tree; degree($v$) delivers the out degree of node $v$; function child($v, i$) delivers the $i^{th}$ child of interior node $v$; uncleOrCousin($l$) delivers a leaf node that is descended from a sibling of the parent of leaf node $l$; function becomesLeaf($v, l$) transforms interior node $v$ into a leaf node labelled as $l$; removeChild($l, v$) removes the leaf node $l$ from the list of children of interior node $v$.

Algorithm HardBreakup

**HardBreakup**($T$)
1   **let** $V \leftarrow$ sortedInteriorNodes($T$)
2   **let** $S \leftarrow \emptyset$
3   **let** $i \leftarrow 0$
4   **while** notRoot($V[i]$) $\vee$ degree($V[i]$) $> 2$ **do**
5       **let** $v \leftarrow V[i]$
6       **let** $c_0 \leftarrow$ child($v, 0$)
7       **if** degree($v$) $= 2$
8       **then let** $c_1 \leftarrow$ child($v, 1$)
9               **let** $c_2 \leftarrow$ uncleOrCousinOf($c_0$)
10              $S \leftarrow S \cup \{\text{triple}(c_0, c_1, c_2)\}$
11              $v \leftarrow$ becomesLeaf($v, c_0$)
12              $i \leftarrow i + 1$
13      **else for** $j \leftarrow 1$ **to** degree($v$) $- 2$ **do**
14          **for** $k \leftarrow j + 1$ **to** degree($v$) $- 1$ **do**
15              **let** $c_1 \leftarrow$ child($v, j$)
16              **let** $c_2 \leftarrow$ child($v, k$)
17              $S \leftarrow S \cup \{\text{fan}(c_0, c_1, c_2)\}$
18          $v \leftarrow$ removeChild($c_0, v$)
19      **return** $S$

Figure 12: Hard breakup of a tree $T$, producing triples and fans.

that when encountering a $k-$fan this is broken up into $\binom{n}{3}$ 3-fans, and in a soft breakup a fan is broken into a linear number of rooted triples. Algorithms for hard and soft breakups are given in Figures 12 and 13, and are used by the imperative OneTree algorithm here and the constraint programming models.

Algorithm SoftBreakup
_______________________________________________

**SoftBreakup**($T$)
1   **let** $V \leftarrow$ sortedInteriorNodes($T$)
2   **let** $S \leftarrow \emptyset$
3   **let** $i \leftarrow 0$
4   **while** notRoot($V[i]$) **do**
5       **let** $v \leftarrow V[i]$
6       **let** $c_0 \leftarrow$ child($v, 0$)
7       **let** $c_1 \leftarrow$ child($v, 1$)
8       **let** $c_2 \leftarrow$ uncleOrCousinOf($c_0$)
9       $S \leftarrow S \cup \{\text{triple}(c_0, c_1, c_2)\}$
10      **if** degree($v$) = 2
11      **then** $v \leftarrow$ becomesLeaf($v, c_0$)
12          $i \leftarrow i + 1$
13      **else** $v \leftarrow$ removeChild($c_0, v$)
14  **return** $S$

Figure 13: Soft breakup of a tree $T$, producing only triples.

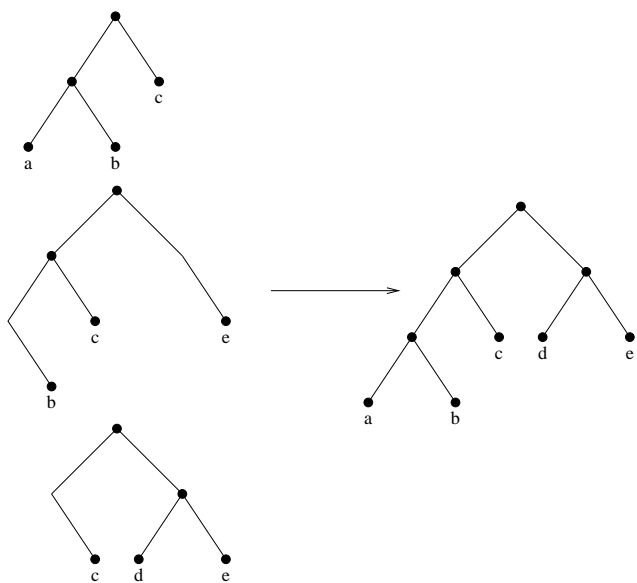

Figure 14: A toy input (left) and single solution to the supertree problem (right). Input
trees are distorted to make relationships in resultant supertree more obvious.

A toy set of input triples and single solution are shown in Figure 14. The triples have
been drawn to reflect how the solution is compatible with them.

Ng and Wormald (1996) give the complexity of OneTree as $O(h(n))$ where $h(n) = n(n + t + bn)\alpha(n + t + f)$, $n$ is the number of labels, $t$ the number of triples, $f$ the number
of fans, $b$ is the sum of the squares of the number of leaves in the fans, and $\alpha$ is the inverse
Ackermann function (and is less than 4 for all conceivable inputs and so behaves like a
constant). Therefore if the input trees are fully resolved (i.e., have no fans) then running time complexity is $O(n^2)$ but in the worst case complexity grows to $O(n^4)$. This should be contrasted with the $O(t \cdot n)$ complexity of Bryant and Steel's OneTree (1995).

## 4.2 A Constraint Encoding using Toolkit Constraints

This second stage, i.e., OneTree equivalent, was first solved as a constraint program by Gent et al. (2003). The encoding takes advantage of an equivalence between *ultrametric trees* and *ultrametric matrices*:

**Definition 7.** *Let $M$ be a real symmetric $n \times n$ matrix. An* ultrametric tree *for $M$ is a rooted tree $T$ such that:*

1. *$T$ has $n$ leaves, each corresponding to a unique row of $M$;*

2. *each internal node of $T$ has at least 2 children;*

3. *for any two leaves $i$ and $j$, $M_{ij}$ is the label of the most recent common ancestor of $i$ and $j$; and*

4. *along any path from the root to a leaf, the labels strictly increase.*

**Theorem 3.** *A symmetric matrix $M$ has an ultrametric tree $T$ if and only if it is an ultrametric matrix. Furthermore, the tree $T$ uniquely determines the matrix $M$ and the matrix $M$ uniquely determines the tree $T$.*

*Proof.* A proof is given by Gusfield (1997). □

There is a clear correspondence between Definition 7 and the description of a species tree given in Section 4: a species tree $T$ is an ultrametric tree for matrix $M$, where $M_{ij}$ is the depth of the mrca of species $i$ and $j$ or $M_{ij}$ is the divergence date of those two species. For this reason we can use an ultrametric matrix model to solve the supertree problem.

### 4.2.1 The Model of Gent et al.

Given as input a forest $F$ with $n$ distinct leaf labels, a symmetric $n \times n$ matrix $M$ of constrained integer variables is created with domains $\{1, \ldots, n-1\}$ or $\{0\}$ on the main diagonal. Variable $M_{ij}$ is the depth of the mrca of species $i$ and $j$. Initially, constraints are posted to make the whole matrix ultrametric thus ensuring that any resulting tree is ultrametric:

$$
\begin{aligned}
& M_{ij} > M_{ik} = M_{jk} \\
\vee \quad & M_{ik} > M_{ij} = M_{jk} \\
\vee \quad & M_{jk} > M_{ij} = M_{ik} \\
\vee \quad & M_{ij} = M_{ik} = M_{jk}
\end{aligned}
\tag{2}
$$

for each $i < j < k$. The input trees are then broken up into triples and fans using either of the breakup algorithms of Figures 12 and 13. For each triple $(ij)k$ produced the constraint

$$
M_{ij} > M_{ik} = M_{jk}
\tag{3}
$$

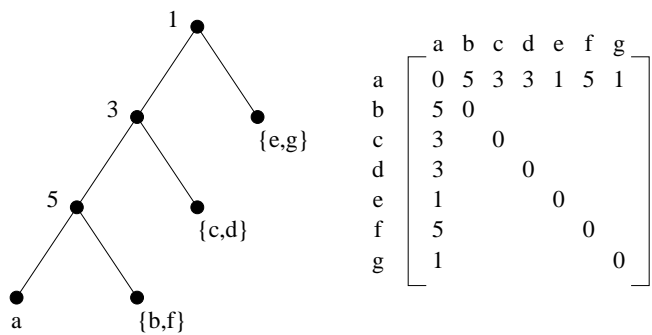

Figure 15: One iteration of an algorithm to convert an ultrametric matrix to a tree

is posted and for each 3-fan $(ijk)$

$$M_{ij} = M_{ik} = M_{jk} \qquad (4)$$

is posted. These constraints break the disjunctions of Equation 2. The model has $(n^2-n)/2$ variables and

$$t + f + \binom{n}{3} = O(n^3) + O(n^3) + \Theta(n^3) = \Theta(n^3) \qquad (5)$$

constraints, where $t$ is the number of triples and $f$ the number of fans. There are $O(n^3)$ of each because each one breaks the disjunction in at most one constraint from Equation 2, and there are $\Theta(n^3)$ of those.

### 4.2.2 Converting Back to Tree Representation

The final step is to use an algorithm based on the constructive proof by Gusfield (1997) of the $\Leftarrow$ direction of Theorem 3 to build a tree from the matrix $M$ produced by a constraint solver. We will not describe this algorithm in detail, but for the sake of intuition it works as follows

- Pick an arbitrary leaf $s$. Let the number of distinct entries in row $s$ be $d$.

- Partition the other leaves into sets $p_1, \ldots, p_d$ based on their entry in row $s$.

- Solve the problem recursively on each $p_i$ by ignoring all the rows and columns in the matrix $D$ not in $p_i$.

- Combine into overall solution by attaching subproblem solutions at the correct depth on the path to $s$.

Figure 15 shows one recursion of the algorithm with a choice of leaf $a$ and shows that row $a$ fully describes the path to $a$ in the corresponding tree.

---

Algorithm CPBuild

---

**CPBuild**$(F)$
1    **let** $(V, D, C) \leftarrow$ CPModel$(F)$
2    **for** $T \in F$ **do**
3        **for** $t \in$ BreakUp$(T)$ **do** $C \leftarrow$ post$(t, C)$
4    **if** propagate(V,D,C) **then return** UMToTree(V,D)
5    **else fail**()

---

Figure 16: Build a supertree from forest $F$ using the ultrametric constraint model.

### 4.2.3 Time Complexity of the Model of Gent et al.

BreakUp, and the procedures to build a constraint model and to convert an ultrametric matrix to a tree are all polynomial time. However the complexity of backtracking search over $O(n^2)$ variables with $O(n)$-size domains is worst case $O(n^{n^2})$. This is an upper bound on the time taken to solve the supertree problem. We have not attempted to derive a lesser upper bound on the time complexity, since, as we will show in the following section, our new model has provably achieved a polynomial time bound.

## 4.3 A Constraint Encoding Using the New Propagator Design

This issue of potentially exponential solution time model of Section 4.2 is worrying, but in our experiments the time taken to solve instances has not been a major issue. Conversely the memory requirements are a problem in practice, but not in theory! The model requires $\Theta(n^3)$ space for $n$ species, but the constant factor is inhibiting. Posting the constraint of Equation 1 literally (using toolkit propagators) as described in Section 3 uses 23 propagators in the JChoco toolkit. This requires roughly 23 times the runtime memory of a single propagator, since each corresponds to a single Java object, and each of these have comparable footprints. As we will show in the empirical study of Section 5, this prevents modest instances from being loaded on typical current workstations.

Using the new propagator of Section 3 we replace these $\binom{n}{3}$ propagators with a single compact propagator and as a result memory usage is reduced asymptotically from $\Theta(n^3)$ to $\Theta(n^2)$ since now the model memory is dominated by the $\Theta(n^2)$ space needed for the matrix $M$. Also reducing the amount of space to be initialised delivers a proportional saving in build time. But most importantly, using the new constraint provides a solution to exponential time complexity, because enforcing BC(Z) on the model allows a solution to be read out of the lower bound of each domain. Theorem 4 is a proof of correctness for this algorithm.

Figure 16 gives a schema for the constraint programming algorithm for supertree construction, CPBuild. The algorithm takes as input a forest $F$ of trees. In line 1 a constraint model is produced, i.e., an $n \times n$ symmetric array of constrained integers variables is created, where there are $n$ unique species in the forest, and the UM-Matrix-BCZ constraint is then posted over those variables. Lines 2 and 3 breaks these input trees into their triples and fans using either of the breakup algorithms given in Figures 12 and 13, and posts them into the model as constraints. Propagators for the constraints are executed to a fixed point in line 4; if this succeeds a tree is created from the lower bounds of the ultrametric matrix otherwise we fail.

**Lemma 7.** *If the propagator for every constraint in a model enforces BC(Z) and further-more the lower bounds are mutually supportive, then after executing all propagators to a fixed point, either the lower bounds are a solution, or we have an empty domain and fail.*

*Proof.* If we reduce each domain to just the lower bound after a fixed point is obtained then each bound is supported because they are mutually supportive by supposition. Hence every constraint is simultaneously satisfied by these singleton domains and, by definition, we have a solution. □

**Theorem 4.** CPBUILD *is a polynomial time solution to the supertree problem.*

*Proof.* The only constraints involved in the model are those for triples and fans and ultra-metric constraints. By Theorem 1 we know that all lower bounds are supported after the propagators run, and by Lemma 1 we know that the lower bounds are mutually support-ive. The same is true of the disjunction-breaking propagators. Hence by Lemma 7, and as shown in Figure 16, we can either read out a solution or fail. We can enforce BC(Z) on the problem in polynomial time as shown below. □

Immediate from this is that we can preserve the polynomial time solution with the addition of a polynomial number of side-constraints, so long as these additional constraints preserve the property that lower bounds are mutually supportive. In fact, CSPs such as these with ordered domains where all constraints have the property that lower bounds are mutually supportive belong to a known tractable class called min-closed (Jeavons & Cooper, 1995).

### 4.3.1 TIME COMPLEXITY OF CPBUILD

The algorithm can be implemented to run in $O(n^4)$ time using a variation on the AC3 algorithm. AC3 (Mackworth, 1977) begins with a queue containing all constraints. It repeatedly removes a constraint until none remain and runs the associated propagator. Any constraints over affected variables are re-queued, if necessary. Once the queue empties, all propagators are at a fixed point. We need $O(n^3)$ constraints so the worst case complexity is

$$\underbrace{O(n^3)}_{\text{build initial Q}} + \underbrace{O(n)O(n^3)}_{\text{worst case re-queues with 1 value removed at a time}} \cdot \underbrace{O(1)}_{\text{propagation time}}$$

or $O(n^4)$ overall. This matches the worst case complexity of ONETREE (Ng & Wormald, 1996). Our constraint solution has its worst case when the problem is unsolvable, since when it is unsolvable domains are *emptied* by propagation, whereas for solvable instances propagation reaches a fixed point sooner.

## 5. Empirical Study

We present an empirical study to determine if any practical improvements have been achieved in constraint solutions to the supertree problem and, if so, what size of improve-ment. Experiments were run using a 1.7GHz Pentium 4 processor with 768MB of memory, using Sun Java build 1.5.0_06-b05. The constraint toolkit used was JCHOCO version 1.1.04.

Input trees were broken up using a hard breakup, consequently in all cases fans were treated as hard polytomies[2].

Our benchmark is real-life seabird data previously used by Kennedy and Page (2002) and Beldiceanu et al. (2008) and we present statistics on various techniques for producing supertrees, namely OneTree, and the CP solutions of Section 4 (entries Toolkit and CPBuild). For completeness we reproduce the results of Beldiceanu et al. (2008) over the same data set, and tabulate this as TreeCon. TreeCon uses a single-successor model, where constrained integer variables represent nodes within a tree, and domains correspond to possible successors[3]. A unique variable represents the root and loops on itself (i.e., $v_{root} = root$), and leaf nodes have an indegree of zero. Precedence and incomparability constraints are then generated from the input trees.

The TreeCon results were encoded in the same constraint programming toolkit as ours but on a processor that was approximately twice as fast (3GHz). We do not correct the times to compensate for this factor. We mark in **bold** results that differ very significantly between the CPBuild and TreeCon results, specifically those whose runtimes would undoubtedly be a factor of 10 different on the same processor. Results are reported for combinations of seabird trees (input trees named A to G) and the following data is tabulated below:

**Data** The combination attempted.

**n** Total distinct species in input trees

**Sol** T iff supertree is possible

**Technique** Type of algorithm used to solve

**Build** Time in milliseconds to initialise CP model

**Solve** Time in milliseconds to first solution, if any

**Total** = Build + Solve

**Nodes** Number of nodes in search tree

**Mem** Model memory in MB

In the table, DNL means that the model could not be loaded (as it was too large) and DNF means that it could not be solved within 30 mins, but succeeded in loading. We have not provided the memory usage of OneTree; however it is smaller than that of any of the constraint encodings.

The most obvious thing to note is how much faster the imperative approach is compared to the constraint techniques. Why is this? Primarily it is due to the lower complexity of OneTree in the absence of fans (we have not investigated if we can benefit from this), and partly due to the generality of the constraint programming approach. The imperative approach is highly specialised to only one class of problem whereas the constraint approach sits within a toolkit, and runs on top of a general purpose constraint maintenance system. We should not expect that the constraint approach will compete in raw speed but what we later demonstrate (in Section 6) is that the approach benefits from its versatility, i.e., the costs of space and time is repaid by the ease of accommodating variants of the problem into the same model.

---

2. In a later section we use soft breakup.

3. An alternative constraint model of a tree might use 0/1 variables corresponding to potential edges within an adjacency matrix (Prosser & Unsworth, 2006), or indeed the CP(Graph) computation domain (Dooms, 2006).

| Data | $n$ | Sol | Technique | Build | Solve | Total | Nodes | Mem |
|------|-----|-----|-----------|-------|-------|-------|-------|-----|
| AB | 23 | T | Toolkit | 2056 | 374 | 2430 | 23 | 26.92 |
| | | | CPBuild | 183 | 131 | 314 | 23 | 0.24 |
| | | | TreeCon | | | 302 | | |
| | | | ONETREE | | | 13 | | |
| AC | 32 | F | Toolkit | 2670 | 327 | 2997 | 0 | 36.34 |
| | | | CPBuild | 189 | 153 | 342 | 0 | 0.34 |
| | | | TreeCon | | | 406 | | |
| | | | ONETREE | | | 12 | | |
| AD | 47 | T | Toolkit | 8235 | 946 | 9181 | 38 | 118.51 |
| | | | CPBuild | 220 | 248 | 468 | 38 | 0.70 |
| | | | TreeCon | | | 398 | | |
| | | | ONETREE | | | 22 | | |
| AE | 95 | F | Toolkit | DNL | DNL | DNL | DNL | > 629 |
| | | | CPBuild | 340 | 1477 | 1817 | 0 | 2.79 |
| | | | TreeCon | | | **10393** | | |
| | | | ONETREE | | | 37 | | |
| AF | 31 | T | Toolkit | 2497 | 379 | 2876 | 19 | 32.99 |
| | | | CPBuild | 188 | 137 | 325 | 18 | 0.32 |
| | | | TreeCon | | | 127 | | |
| | | | ONETREE | | | 20 | | |
| AG | 46 | T | Toolkit | 7671 | 871 | 8542 | 31 | 111.07 |
| | | | CPBuild | 222 | 252 | 474 | 31 | 0.68 |
| | | | TreeCon | | | 409 | | |
| | | | ONETREE | | | 21 | | |
| BC | 29 | F | Toolkit | 2056 | 21931 | 23987 | 171 | 26.90 |
| | | | CPBuild | 171 | 107 | 278 | 0 | 0.27 |
| | | | TreeCon | | | 32 | | |
| | | | ONETREE | | | 8 | | |
| BD | 42 | T | Toolkit | 5833 | 930 | 6763 | 33 | 84.26 |
| | | | CPBuild | 201 | 251 | 452 | 33 | 0.55 |
| | | | TreeCon | | | 301 | | |
| | | | ONETREE | | | 17 | | |
| BE | 94 | F | Toolkit | DNL | DNL | DNL | DNL | > 629 |
| | | | CPBuild | 335 | 16340 | **16675** | 0 | 2.71 |
| | | | TreeCon | | | 892 | | |
| | | | ONETREE | | | 11 | | |
| BF | 30 | T | Toolkit | 2405 | 343 | 2748 | 29 | 29.83 |
| | | | CPBuild | 174 | 99 | 273 | 29 | 0.28 |
| | | | TreeCon | | | 144 | | |
| | | | ONETREE | | | 8 | | |
| BG | 40 | T | Toolkit | 5098 | 651 | 5749 | 30 | 72.71 |
| | | | CPBuild | 203 | 353 | 556 | 30 | 0.51 |
| | | | TreeCon | | | 1440 | | |
| | | | ONETREE | | | 13 | | |
| CD | 45 | T | Toolkit | 10056 | 1134 | 11190 | 45 | 143.91 |
| | | | CPBuild | 224 | 276 | 500 | 45 | 0.77 |
| | | | TreeCon | | | 630 | | |
| | | | ONETREE | | | 14 | | |
| CE | 68 | T | Toolkit | DNL | DNL | DNL | DNL | > 629 |
| | | | CPBuild | 516 | 1451 | 1967 | 68 | 2.72 |
| | | | TreeCon | | | **27180** | | |
| | | | ONETREE | | | 36 | | |
| CF | 34 | T | Toolkit | 3101 | 563 | 3662 | 30 | 43.72 |
| | | | CPBuild | 180 | 133 | 313 | 30 | 0.36 |
| | | | TreeCon | | | 393 | | |
| | | | ONETREE | | | 11 | | |
| CG | 44 | F | Toolkit | 6683 | 587 | 7270 | 0 | 97.10 |
| | | | CPBuild | 210 | 215 | 425 | 0 | 0.61 |
| | | | TreeCon | | | 1530 | | |
| | | | OneTree | | | 14 | | |
| DE | 104 | F | Toolkit | DNL | DNL | DNL | DNL | > 629 |
| | | | CPBuild | 360 | 2021 | 2381 | 0 | 3.31 |
| | | | TreeCon | | | 1126 | | |
| | | | OneTree | | | 34 | | |
| DF | 44 | T | Toolkit | 6613 | 987 | 7600 | 37 | 97.10 |
| | | | CPBuild | 203 | 250 | 453 | 37 | 0.60 |
| | | | TreeCon | | | 630 | | |
| | | | OneTree | | | 17 | | |
| DG | 56 | F | Toolkit | 14090 | 2280 | 16370 | 1 | 201.42 |
| | | | CPBuild | 252 | 640 | 892 | 0 | 0.99 |
| | | | TreeCon | | | 910 | | |
| | | | OneTree | | | 19 | | |
| EF | 94 | F | Toolkit | DNL | DNL | DNL | DNL | > 629 |
| | | | CPBuild | 331 | 9546 | 9877 | 0 | 2.71 |
| | | | TreeCon | | | 1035 | | |
| | | | OneTree | | | 12 | | |
| EG | 97 | F | Toolkit | DNL | DNL | DNL | DNL | > 629 |
| | | | CPBuild | 344 | 8900 | 9244 | 0 | 2.89 |
| | | | TreeCon | | | 1211 | | |
| | | | OneTree | | | 15 | | |
| FG | 38 | F | Toolkit | 4299 | DNL | DNL | DNL | 61.41 |
| | | | CPBuild | 195 | 212 | 407 | 0 | 0.46 |
| | | | TreeCon | | | 62 | | |
| | | | OneTree | | | 10 | | |
| ABDF | 72 | T | Toolkit | 27032 | 5291 | 32323 | 63 | 382.52 |
| | | | CPBuild | 277 | 722 | 999 | 59 | 1.48 |
| | | | TreeCon | | | **8139** | | |
| | | | OneTree | | | 34 | | |
| ABDG | 78 | F | Toolkit | 60847 | DNF | DNF | DNF | 553.49 |
| | | | CPBuild | 301 | 3633 | 3934 | 0 | 1.91 |
| | | | TreeCon | | | 347 | | |
| | | | OneTree | | | 29 | | |
| ACDF | 72 | F | Toolkit | 31067 | 1931 | 32998 | 0 | 434.84 |
| | | | CPBuild | 286 | 649 | 935 | 0 | 1.61 |
| | | | TreeCon | | | **8690** | | |
| | | | OneTree | | | 28 | | |
| ACDG | 81 | F | Toolkit | DNL | DNL | DNL | DNL | > 629 |
| | | | CPBuild | 307 | 1711 | 2018 | 0 | 2.06 |
| | | | TreeCon | | | **12650** | | |
| | | | OneTree | | | 35 | | |
| ACE | 97 | F | Toolkit | DNL | DNL | DNL | DNL | > 629 |
| | | | CPBuild | 737 | 1632 | 2369 | 0 | 2.91 |
| | | | OneTree | | | 38 | | |

The most impressive aspect of the matrix model of Section 4.3 over that of Section 4.2 is the improvement in memory requirements, so that all instances can now be loaded comfortably. This also has a dramatic impact on the build time. These improvements dominate the reduction in solve time in practice. The toolkit model is outperformed by CPBuild by an order of magnitude on each instance; moreover, there are two cases of search occurring in the toolkit model (on data sets BC and ABDG) whereas CPBuild never has to search. The polynomial time complexity is due to the provable absence of search.

Our results also compare well against those of Beldiceanu et al. (2008). There is one case, BE, where CPBuild is an order of magnitude slower than TreeCon; so far we do not have an explanation for this. There are four cases when TreeCon is significantly worse than CPBuild. No results are available for TreeCon over the data set ACE. It should be noted that Beldiceanu et al. do not yet have a complete filtering algorithm for this problem based on their constraint model and, from personal communication, although the TreeCon model never backtracked over the birds data set there is as yet no proof that the complexity of their model is polynomial. It should also be noted that we see CPBuild taking more time on unsolvable instances than solvable instances, as predicted.

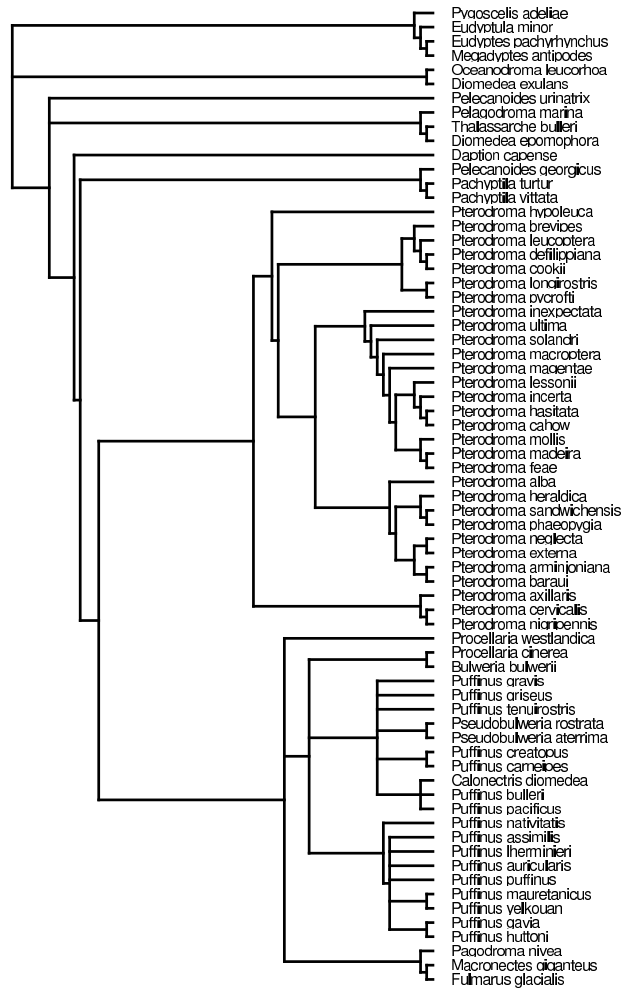

Figure 17: Supertree with largest compatible data sets of birds ABDF. This took 737ms to model and 270ms to solve using CPBUILD.

Figure 17 shows the supertree, displayed with treeView (Page, 1996), produced from the largest compatible forest $\{A, B, D, F\}$. This supertree has 72 leaves and takes about 1 second to produce. Although the result is not printed in the table, finding the forest $\{A, B, C, D, E, F, G\}$ incompatible takes about 12 seconds in total (1.5 seconds to build the model and 10 seconds to determine incompatibility).

## 6. Versatility of the Constraint Model

One of the strengths of constraint programming is its versatility: given a constraint model of a core problem this model can then be enhanced to address variants of the original pure problem. We demonstrate this versatility with respect to the ultrametric model, presenting four variants of the supertree problem (a) incorporating ancestral divergence dates into the model, (b) nested taxa, (c) determining if an induced triple or fan is common to all supertrees, and (d) coping with incompatibilities.

### 6.1 Ancestral Divergence Dates

Semple et al. (2004) and Bryant et al. (2004) add temporal information to the input trees. Interior nodes may be labelled with integer *ranks* such that if interior node $v_2$ is a proper descendant of $v_1$ then $rank(v_1) < rank(v_2)$, resulting in a ranked phylogenetic tree. Additionally relative divergence dates may be expressed in the form "div(c,d) *predates* div(a,b)" and this is interpreted as "the divergence of species $c$ and $d$ predates that of species $a$ and $b$". The RANKEDTREE algorithm (Bryant et al., 2004) takes as input a collection of precedence constraints derived from input ranked species trees and *predates* relations. The algorithm outputs a ranked tree that respects those relations or returns "not compatible".

This is trivial to incorporate into the constraint model. If trees have been ranked then for each pair of species $(i, j)$ in the leaf set we instantiate the constrained integer variable $M_{ij}$ to the value of $mrca(i, j)$. For a *predates* relation "div(c,d) *predates* div(a,b)" we post the constraint $M_{cd} < M_{ab}$. This is done before step 4 of CPBUILD (Figure 16), i.e., ranks and *predates* relations become side constraints. Similarly time bounds on speciation events are posted as unary constraints, i.e. a *dated phylogenetic tree* upper and lower divergence bounds are given on interior nodes, such that $l(a, b)$ and $u(a, b)$ give respectively the lower and upper bounds on the divergence dates of species $a$ and $b$. In the constraint program the following two side constraints are then posted (again, before step 4): $l(a, b) \leq M_{ab}$ and $M_{ab} \leq u(a, b)$.

A demonstration of ranked trees is given in Figure 18. On the left we have two ranked species trees of cats used recently by Semple et al. (2004) and originally by Janczewski, Modi, Stephens, and O'Brien (1995). The branch lengths of the source trees have been translated into rankings and added to the interior vertices of those trees. On the right we have one of the 17 possible resultant supertrees. In total, 7 of the 17 solutions contain interior nodes with ranges. If interior nodes are labelled with specific values rather than ranges then 30 solutions are produced, some of which are structurally identical. This goes some way to addressing the issue of enumerating all supertrees compactly, raised as a challenge by Semple et al. (2004). In Figure 19 we show the effect of adding a *predates* constraint to a supertree construction. The data has previously been used by Bryant et al. (2004) in their Figures 5 and 6.

### 6.2 Nested Taxa

A *taxon* (plural, *taxa*) is a group of organisms comprising a single common ancestor and its descendents (Dawkins & Wong, 2004). For example the species "lion" and the class "birds" are taxa. So far, all our species trees have been leaf-labelled, however this is restrictive because trees may be annotated with taxa names on both leaves and internal nodes, giving *nested taxa*. For example, Figure 20 shows tree $T_1$ with an internal node labelled $P$ which has descendents $a$ and $b$, i.e., the $a$ and $b$ taxa are nested within the $P$ taxon. Problems related to creating compatible supertrees for this type of data were raised by Page (2004) and defined and solved by Daniel and Semple (2004). A set of input trees and possible solution to the problem are shown in Figure 20: notice that all labels are conserved in the solution, all ancestral relationships are conserved and, for any labels $l_i$ and $l_j$ from the same input tree, $l_i$ is ancestor of $l_j$ in the input tree if and only if $l_i$ is ancestor of $l_j$ in the solution tree. This is an instance of the problem *Higher Taxa Compatibility* defined by Daniel and Semple (2004) and Semple et al. (2004), where the result tree must *perfectly display* all of the input trees. We now define the problem more formally.

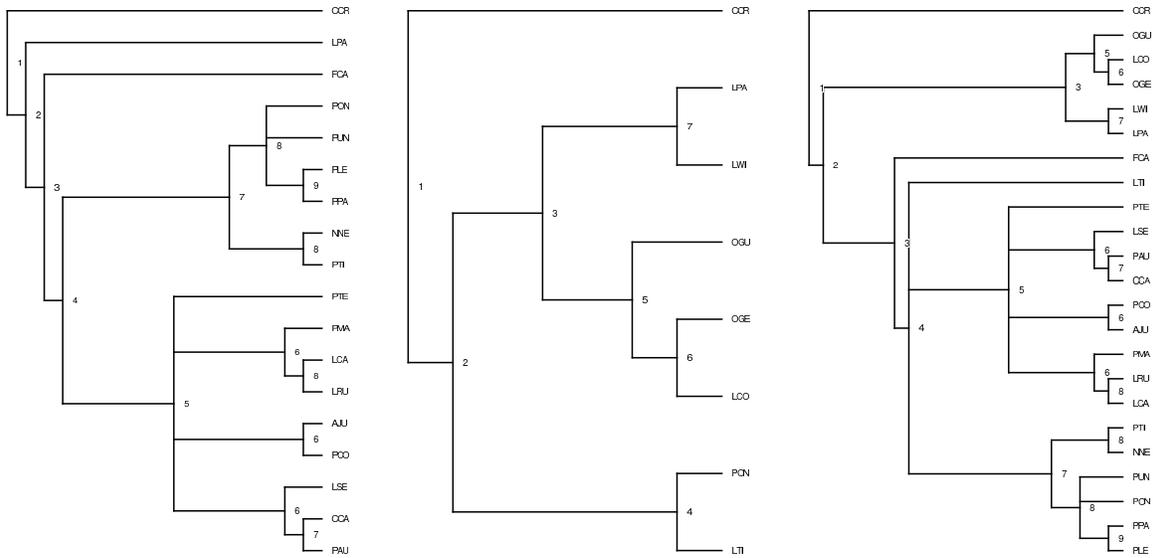

Figure 18: Two ranked trees of cats. On the right one of the 17 possible supertrees produced by CPBUILD. Displayed using Page's treeView.

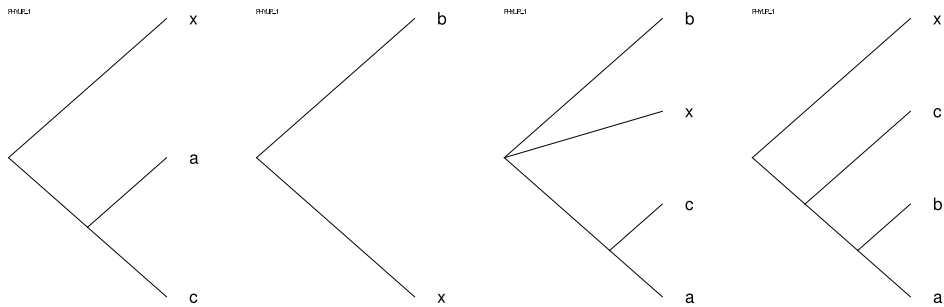

Figure 19: Two input trees $T_1 = ((a, c), x)$ and $T_2 = (b, x)$ with a resultant supertree shown in the 3rd position. The tree on the far right is a supertree from $T_1$ and $T_2$ with the side constraint "div(a,c) *predates* div(a,b)", produced by CPBuild and displayed using Page's treeView.

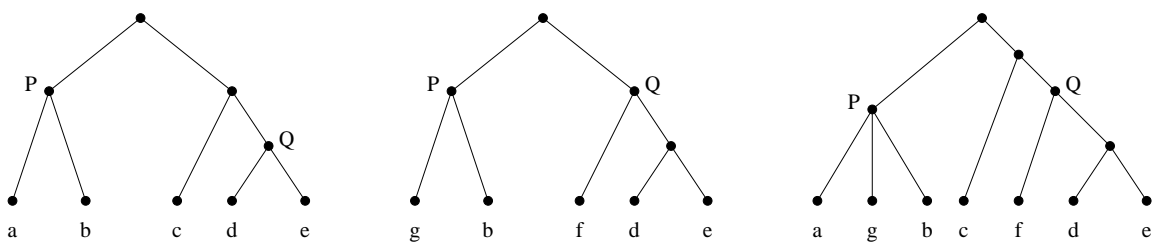

Figure 20: Two input rooted $X$-trees $T_1$ and $T_2$ (left) and an output tree $T_3$ (right) that perfectly displays them.

**Definition 8.** *A* rooted $X$-tree *(Daniel & Semple, 2004) is a species tree where internal nodes as well as leaves may be labelled from the set $X$.*

In the following we will be slightly loose and may use a label $l$ to identify the labelled node, as well as the label itself, e.g., "descendants of $l$" means the same as "descendants of the node labelled $l$".

**Definition 9.** *A rooted $X$-tree $T$ perfectly displays a rooted $X'$-tree $T'$ when*

1. $X' \subseteq X$;

2. $T'$ displays $T$, neglecting internal labels;

3. *if $a$ is a descendant of $b$ in $T'$ then $a$ is a descendant of $b$ in $T$; and*

4. *if $a$ is not a descendant of $b$ in $T'$ then $a$ is not a descendant of $b$ in $T$.*

*A rooted $X$-tree $T$ perfectly displays a forest of phylogenetic trees $F = \{T_1, \ldots, T_n\}$ when it perfectly displays every $T_i$.*

### 6.2.1 CONSTRAINT ENCODING

Our constraint encoding is implemented by the addition of variables and side constraints to the standard model of Section 4. We describe how to transform the input to make the constraint solution simpler, and then describe the variables and constraints needed.

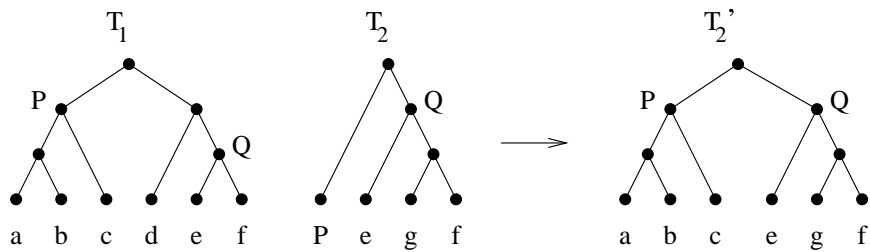

Figure 21: Two input trees $T_1$ and $T_2$ with an enclosing taxon $P$. By a process of substitution $T_2$ is replaced with $T_2'$.

**Substitution**   Taxon $P$ in Figure 21 appears on an internal node of $T_1$, we will call such a label an *enclosing taxon*. Note also that it appears *on a leaf* in $T_2$. The input trees are preprocessed to replace any tree with an enclosing taxon $P$ on a leaf by the same tree with any single subtree rooted at $P$ substituted in its place. There must be such a subtree elsewhere in the input forest, or we have a contradiction that $M$ is an enclosing taxon. This process does not add or remove any information, since the relationships between $M$ and everything in the tree still holds, and the new relationships between the taxa in the subtree at $M$ and the rest of the tree were always implicit in the input.

The aim of this process was just to obtain a set of inputs where enclosing taxa appear on internal nodes only, because without loss of generality our constraint encoding assumes that this is the case. Figure 21 shows an example of the substitution process applied to trees $T_1$ and $T_2$, and $T_2$ would be replaced with $T_2'$.

**Variables and constraints**   The variables added are one integer variable $v_l$ per enclosing taxa/label $l$, each with a domain of $\{1, \ldots, n-1\}$. The value of this variable in a solution is the tree depth of the internal node which it labels, and the label's position in the final tree is determined because it must be at the unique node of that depth on a path from one of its nested taxa to the root. See Figure 15 and suppose for the sake of argument that we have an enclosing taxa $M$ which labels b, and the variable $l_M = 1$ in a solution. The unique location where the label $M$ can go is at the root node.

Properties (1) and (2) from Definition 9 above are immediate from the properties of the earlier model which is the foundation for this one. Before explaining how to enforce property (3) we introduce some notations for convenience. Function $desc(l, F)$ returns the set of all descendants of label $l$ in any tree $T$ in the forest $F$, and $notDesc(l, T)$ returns the set of labels that are not descendants of $l$ in tree $T$.

We first need a constraint that $l$ must label every single species that it labels in an input. For every enclosing label $l$, post the following set of constraints:

$$\{v_l \leq M_{ij} \mid i \in desc(l, F) \ \wedge \ j \in desc(l, F) \ \wedge \ i \neq j\} \tag{6}$$

so that the label must settle at least as shallow as any mrca of its descendants in input, and hence they must remain descendants. Notice that it is necessary to consider pairs of species from distinct input trees. An alternative of taking pairs from the same tree does not work, because it is necessary for all pairs to be under the *same* internal node $l$, rather than two distinct nodes that happen to be at the correct depth. Next, the label must be constrained so that no label that was not already a descendant becomes one. For each $X$-tree $T$ and enclosing label $l \in X$, post the following set of constraints:

$$\{v_l > M_{ij} \mid i \in desc(l, \{T\}) \ \wedge \ j \in notDesc(l, T)\} \tag{7}$$

so that the label $l$ must be placed strictly deeper than any mrca of a descendant and something that's not a descendant, i.e., no non-descendents of $l$ can be a descendent in the result. As an illustration we list the generated constraints for the example of Figure 20.

1. Equation 6 and $l = P$: $\{v_P \leq M_{ab}, v_P \leq M_{ag}, v_P \leq M_{bg}\}$

2. Equation 6 and $l = Q$: $\{v_Q \leq M_{de}, v_Q \leq M_{df}, v_Q \leq M_{ef}\}$

3. Equation 7, $l = P$ and $T \in \{T_1, T_2\}$: $\{v_P > M_{ac}, v_P > M_{ad}, v_P > M_{ae}, v_P > M_{bc}, v_P > M_{bd}, v_P > M_{be}, v_P > M_{gf}, v_P > M_{gd}, v_P > M_{ge}, v_P > M_{bf}\}$

4. Equation 7, $l = Q$ and $T \in \{T_1, T_2\}$: $\{v_Q > M_{ad}, v_Q > M_{ae}, v_Q > M_{bd}, v_Q > M_{be}, v_P > M_{cd}, v_Q > M_{ce}, v_Q > M_{gf}, v_P > M_{bf}, v_P > M_{gd}, v_P > M_{ge}\}$

The number of new constraints created by both Equations 6 and 7 is bounded by the number of distinct pairs of species, i.e. $\Theta(n^2)$ new constraints.

## 6.3 Necessity

There may be many possible supertrees for a given input forest. One question is then, what relationships are common to all supertrees? The problem of determining if a derived induced triple (or fan) in a supertree is necessary (i.e., common to all possible supertrees) is introduced by (Daniel, 2003) along with the polynomial time decision procedure NECESSITY.

The algorithm NECESSITY in Figure 22 takes as arguments a forest $F$ of trees, assumed to be compatible, and a rooted triple or fan $\tau$ and determines if $\tau$ occurs in every supertree that displays the trees in $F$. The algorithm is a simple modification of CPBUILD, where lines 1 to 3 are essentially the same. In line 4 the negation of the triple $\tau$ is posted to the problem, where $\neg\tau$ is posted as

$$M_{ik} \neq M_{jk} \vee M_{ij} \leq M_{ik} \vee M_{ij} \leq M_{ik} \tag{8}$$

when $\tau = (ij)k$ and posted as

$$M_{ij} \neq M_{ik} \vee M_{ij} \neq M_{jk} \vee M_{ik} \neq M_{jk} \tag{9}$$

when $\tau = (ijk)$. A call is then made to propagate to make the problem arc-consistent (line 5), and if this fails then $\tau$ is necessary, otherwise it is not necessary. The algorithm has the same complexity as CPBUILD.

## 6.4 Coping with Conflict

When a supertree cannot be produced from a pair of trees some of the input triples and fans must be in conflict with one another, either directly or indirectly. Junker's QUICKXPLAIN method (Junker, 2004) discovers a minimal subset of constraints that when posted and propagated result in a failure. This set is not necessarily the smallest possible set but is

```
Algorithm Necessity
────────────────────────────────────────────────
Necessity(F, τ)
1    let (V, D, C) ← CPModel(F)
2    for T ∈ F do
3        for t ∈ BreakUp(T) do C ← post(t, C)
4    C ← post(¬τ, C)
5    return ¬propagate(V, D, C)
```

Figure 22: Does the triple/fan $\tau$ occur in every supertree that displays the trees in $F$?

minimal in the sense that the removal of any element from this set will not constitute a sound explanation, and the addition of any constraint would be redundant. When the set of constraints are input triples and fans, this minimal set is semantically a collection of input data that is incompatible. Junker (2004) state that this method can be achieved by a worst case of $2k \cdot \log_2(n/k) + 2k$ propagations[4], where $k$ is the size of the minimal explanation found and $n$ is the number of constraints.

An alternative approach is to satisfy as many of the input triples and fans as is possible within a reasonable amount of time, i.e., polynomial time. Semple and Steel propose such an algorithm, MinCutSupertree (2000), and this has been refined by Page (2002). We now propose a similar scheme within the constraint programming framework. We call this algorithm GreedyBuild and it works as follows. We associate a constrained integer variable $x$, with a domain of $\{0, 1\}$, to each triple and fan. If the variable is assigned the value 0 then the triple (or fan) is respected, otherwise it is ignored. Therefore for a triple $(ij)k$ we post the constraint of equation 10 and for a 3-fan $(ijk)$ the constraint of equation 11.

$$(x = 0 \land M_{ij} > M_{ik} = M_{jk}) \lor (x = 1 \land (M_{ik} \neq M_{jk} \lor M_{ij} \leq M_{ik} \lor M_{ij} \leq M_{ik})) \quad (10)$$

$$(x = 0 \land M_{ij} = M_{ik} = M_{jk}) \lor (x = 1 \land (M_{ij} \neq M_{ik} \lor M_{ij} \neq M_{jk} \lor M_{ik} \neq M_{jk})) \quad (11)$$

GreedyBuild then instantiates in turn each of the $x$ variables, i.e. the decision variables, preferring the value 0 to the value 1, and after each instantiation the problem is made arc-consistent. The algorithm is shown in Figure 23. In line 1 a constraint model is produced, i.e., an $n \times n$ symmetric array of constrained integers variables is created, where there are $n$ unique species in the forest, and the UM-Matrix-BCZ constraint is then posted over those variables. The variable $X$ of line 2 is then the set of decision variables. The input trees are broken up as before, and a new variable $x$ is created for each triple or fan. In line 6 the constraints of equations 10 and 11 are posted into the model. The loop of lines 10 to 12 in turn select a decision variable, set it to its lowest possible value, and then make the problem arc-consistent. This might in turn cause uninstantiated variables to have the value 0 removed from their domain if their associated triple or fan conflicts with the triple

---

Algorithm GREEDYBUILD

---

**GreedyBuild**$(F)$
1    **let** $(V, D, C) \leftarrow \text{CPModel}(F)$
2    **let** $X \leftarrow \emptyset$
3    **for** $T \in F$ **do**
4        **for** $t \in \text{BreakUp}(T)$ **do**
5            **let** $x \leftarrow newVar(0, 1)$
6            **let** $c \leftarrow \text{newConstraint}((t \wedge x = 0) \vee (\neg t \wedge x = 1))$
7            $X \leftarrow X \cup \{x\}$
8            $V \leftarrow V \cup \{x\}$
9            $C \leftarrow post(c, C)$
10   **for** $x \in X$ **do**
11       $\text{instantiate}(x)$
12       $\text{propagate}(V, D, C)$
13   **return** $\text{UMToTree}(V, D)$

---

Figure 23: Greedily Build a supertree from forest $F$ using the ultrametric constraint model.

or fan that has just been enforced. This process terminates without failure, because any conflicting triples or fans are essentially ignored. In line 13 the ultrametric matrix is then converted to a tree. The complexity of GREEDYBUILD is then $O((t + f) \cdot n^4)$ where there are $t$ triples and $f$ fans.

GREEDYBUILD was applied to the forest of bird data $\{A, B, C, D, E, F, G\}$ from section 5, using a soft breakup. This data is incompatible when we use CPBUILD, however GREEDYBUILD produces the supertree in Figure 24. This supertree contains 121 species. SOFTBREAKUP produced 201 triples, and of those 17 were rejected. It took less than 2 seconds to build the model and about 100 seconds to solve that model. This should be compared to CPBUILD over the same data set, taking 1.5 seconds to build the model and 10 seconds to determine incompatibility. GREEDYBUILD was also applied to the data set ABDF, producing the identical supertree to CPBUILD, in comparable time (890ms to build the model and 578ms to solve).

Having executed GREEDYBUILD the decision variables in the set $X$ (lines 2, 7, 10 and 11) can be analysed to identify the set of triples and fans that have been excluded from the supertree, i.e., if an $x$ variable has been instantiated with the value 1 then its corresponding triple or fan has been ignored.

Note that we do not claim any biological significance in the arbitrary order we use to suppress triples. GREEDYBUILD could be amended to follow the order of MINCUT-SUPERTREE but we have not investigated this. GREEDYBUILD can also be enhanced as follows. Currently if a triple or fan exists in multiple input trees then it occurs only once as a constraint. This information could be exploited by weighting the decision variables to take into consideration the relative weight of evidence for a triple or fan, e.g., the number of times that a triple or fan occurs as input. The decision variables can then be instantiated in non-increasing order of weight, i.e., a variable ordering heuristic can be used. In the extreme GREEDYBUILD can be modified to become OPTBUILD where a full backtracking search is performed with the objective of minimising the sum of the decision variables, but at a potentially exponential cost in time. This would return the tree with the fewest possible input triples suppressed.

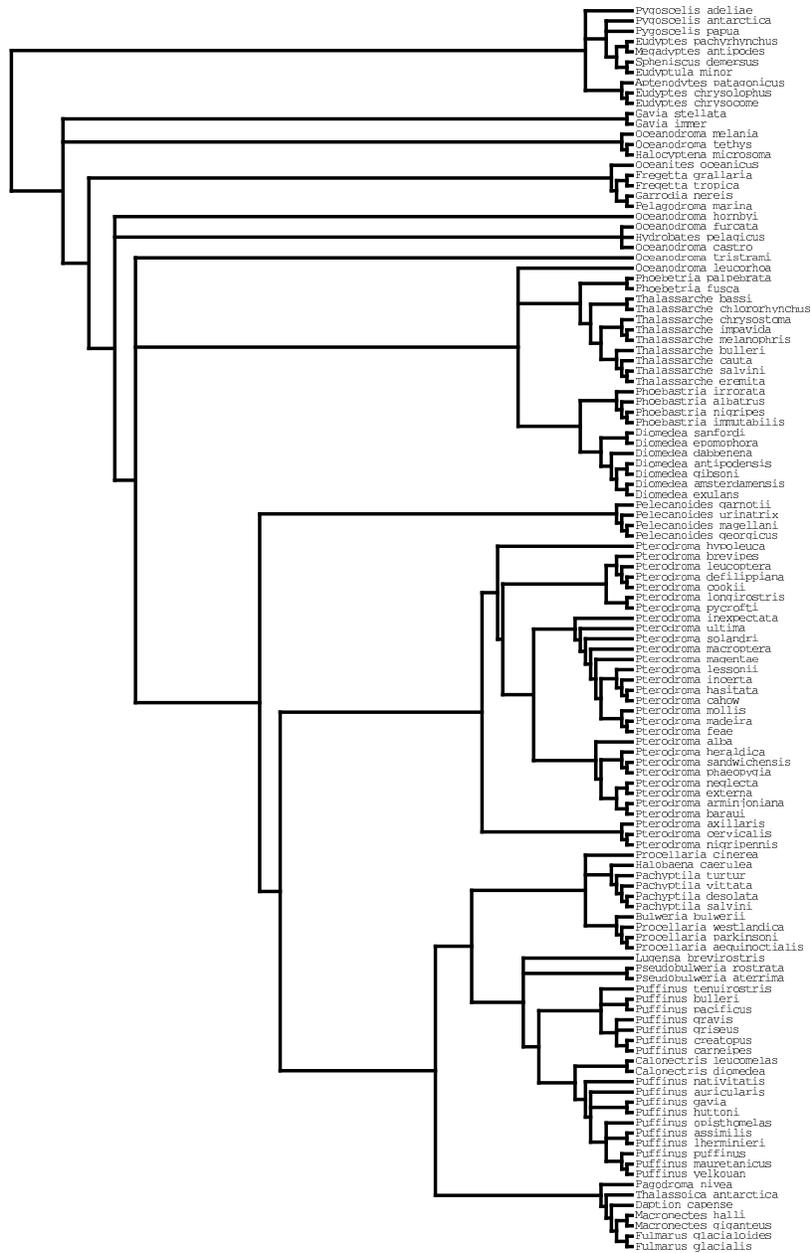

Figure 24: Supertree with largest data set of birds, ABCDEFG, with 121 species. This took about 2 seconds to model and 100 seconds to solve using GREEDYBUILD. Displayed using Rod Page's treeView.

## 6.5 Summary

With little effort, the constraint model has been adapted to deal with ancestral divergence dates and nested taxa. Both have been achieved by adding side constraints. This has an added advantage with respect to ancestral divergence as it can result in a more compact enumeration of output trees when interior nodes are labelled with ranges rather than specific values.

When input trees conflict we propose two options: use QUICKXPLAIN to determine the cause of that conflict or greedily build a supertree using GREEDYBUILD. Bryant et al. (2004) state that they have essentially an "all-or-nothing" approach to supertree construction when using RANKEDTREE and what is needed is something akin to MINCUTSUPERTREE, i.e. when trees are incompatible build a supertree that violates the minimum number of triples or fans, and do this in polynomial time (Page, 2002; Semple & Steel, 2000). This has since been done by Bordewich et al. (2006) and can also be done in our constraint model by incorporating the constraints identified in section 6.1 into GREEDYBUILD.

Although not shown, it is obvious that ancestral divergence data and nested taxa can be combined in the one model, simply by adding all the necessary constraint and auxiliary variables for both variants into the one model. This, again, could be done in GREEDYBUILD, but would require some heuristic or rule to be used when deciding what constraints to ignore when the input trees and side constraints are incompatible.

In our opinion, deriving, combining and analysing the results of imperative algorithms for supertree problems is much more difficult than the above. Most algorithms for variants required far more complex data structures and tailored algorithms for processing them. Moreover, to combine the algorithms once produced seems practically impossible on account of their intricacy. Finally constraint programming provides various generic methods like QUICKXPLAIN "out of the box" that turn out to be of interest in supertree problems.

## 7. Conclusion

We have presented a new constraint propagator for the ultrametric constraint over three integer variables, and shown how this can be extended to a symmetric matrix of constrained integer variables. When bounds(Z)-consistency is established on the symmetric array the lower bounds of variables give mutual support. This is sufficient for modelling and solving the supertree construction problem in $O(n^4)$ time and $O(n^2)$ space, comparable to the complexity of ONETREE (Ng & Wormald, 1996) but inferior to that of the algorithm of Bryant and Steel (1995). So, why bother with the CPBUILD approach when efficient imperative approaches already exist? The answer lies in the versatility of constraint programming. Rather than develop a new algorithm for a new variant of the supertree problem we add side constraints to a base model, and we have shown that a polynomial time bound can often be achieved. We have done this for ancestral divergence dates and nested taxa, we have shown how our model can be used to deliver necessary triples and fans, and we have proposed GREEDYBUILD as a way of dealing with incompatible trees.

## Acknowledgments

We would like to thank Pierre Flener and Xavier Lorca; Barbara Smith, Ian Gent and Christine Wei Wu; Charles Semple, Mike Steel and Rod Page; Muffy Calder and Joe Sventek; Stanislav Zivny; Chris Unsworth; and our three anonymous reviewers/co-authors.

## References

Aho, A., Sagiv, Y., Szymanski, T., & Ullman, J. (1981). Inferring a tree from lowest common ancestors with an application to the optimization of relational expressions. *SIAM J. Comput*, *10*(3), 405–421.

Beldiceanu, N., Flener, P., & Lorca, X. (2008). Combining tree partitioning, precedence, and incompatibility constraints. *Constraints*, *13*, 1–31.

Bessière, C., & Régin, J.-C. (2001). Refining the basic constraint propagation algorithm. In *IJCAI*, pp. 309–315.

Bessière, C. (2006). Constraint propagation. In *Handbook of constraint programming*. Elsevier. Chapter 3.

Bininda-Emonds, O. (2004). *Phylogenetic Supertrees: Combining information to reveal the tree of life*. Springer.

Bordewich, M., Evans, G., & Semple, C. (2006). Extending the limits of supertree methods. *Annals of combinatorics*, *10*, 31–51.

Bryant, D., & Steel, M. (1995). Extension Operations on Sets of Leaf-labeled Trees. *Advances in Applied Mathematics*, *16*, 425–453.

Bryant, D., Semple, C., & Steel, M. (2004). Supertree methods for ancestral divergence dates and other applications. In Bininda-Emonds, O. (Ed.), *Phylogenetic Supertrees: Combining information to reveal the tree of life*, pp. 151–171. Computational Biology Series Kluwer.

Carlier, J., & Pinson, E. (1994). Adjustment of heads and tails for the jobshop scheduling problem. *European Journal of Operational Research*, *78*, 146–161.

Caseau, Y., & Laburthe, F. (1997). Solving small TSPs with constraints. In *Proceedings International Conference on Logic Programming*, pp. 1–15.

Choco (2008). http://www.choco-constraints.net/ home of the choco constraint programming system..

Daniel, P. (2003). Supertree methods: Some new approaches. Master's thesis, Department of Mathematics and Statistics, University of Canterbury.

Daniel, P., & Semple, C. (2004). Supertree algorithms for nested taxa. In Bininda-Emonds, O. (Ed.), *Phylogenetic Supertrees: Combining information to reveal the tree of life*, pp. 151–171. Computational Biology Series Kluwer.

Dawkins, R., & Wong, Y. (2004). *The Ancestor's Tale*. Weidenfeld and Nicholson.

Debruyne, R., & Bessière, C. (1997). Some practicable filtering techniques for the constraint satisfaction problem. In *Proceedings of IJCAI'97*, pp. 412–417.

Dooms, G. (2006). *The CP(Graph) Computation Domain in Constraint Programming*. Ph.D. thesis, Université catholique de Louvain, Faculté des sciences appliquées.

Gent, I., Prosser, P., Smith, B., & Wei, W. (2003). Supertree construction with constraint programming. In *Principles and Practice of Constraint Programming*, pp. 837–841. Springer.

Gusfield, D. (1997). *Algorithms on strings, trees, and sequences: computer science and computational biology*. Cambridge University Press, New York, NY, USA.

Janczewski, D., Modi, W., Stephens, J., & O'Brien, S. (1995). Molecular evolution of mitochondrial 12S RNA and Cytochrome b sequences in the pantherine lineage of Felidae. *Mol. Biol. Evol.*, *12*, 690–707.

Jeavons, P. G., & Cooper, M. C. (1995). Tractable constraints on ordered domains. *Artif. Intell.*, *79*(2), 327–339.

Junker, U. (2004). QUICKXPLAIN: Preferred Explanations and Relaxations for Over-Constrained Problems. In *Proceedings AAAI2004*, pp. 167–172.

Kennedy, M., & Page, R. (2002). Seabird supertrees: Combining partial estimates of procellariiform phylogeny. *The Auk*, *69*, 88–108.

Lhomme, O. (2003). An efficient filtering algorithm for disjunction of constraints. In *Principles and Practice of Constraint Programming*, pp. 904–908. Springer.

Mace, G. M., Gittleman, J. L., & Purvis, A. (2003). Preserving the Tree of Life. *Science*, *300*, 1707–1709.

Mackworth, A. (1977). Consistency in networks of relations. *Artificial Intelligence*, *8*, 99–118.

Ng, M. P., & Wormald, N. C. (1996). Reconstruction of rooted trees from subtrees. *Discrete Appl. Math.*, *69*(1-2), 19–31.

Page, R. (1996). TREEVIEW: An application to display phylogenetic trees on personal computers. *Computer Applications in the Biosciences*, *12*, 357–358.

Page, R. (2004). Taxonomy, supertrees, and the tree of life. In Bininda-Emonds, O. (Ed.), *Phylogenetic Supertrees: Combining information to reveal the tree of life*, pp. 247–265. Computational Biology Series Kluwer.

Page, R. D. M. (2002). Modified mincut supertrees. In *WABI '02: Proceedings of the Second International Workshop on Algorithms in Bioinformatics*, pp. 537–552 London, UK. Springer-Verlag.

Pennisi, E. (2003). Modernizing the Tree of Life. *Science*, *300*, 1692–1697.

Prosser, P. (2006). Supertree construction with constraint programming: recent progress and new challenges. In *WCB06 - Workshop on Constraint Based Methods for Bioinformatics*, pp. 75–82.

Prosser, P., & Unsworth, C. (2006). Rooted Tree and Spanning Tree Constraints. In *17th ECAI Workshop on Modelling and Solving Problems with Constraints*.

Régin, J.-C. (1994). A filtering algorithm for constraints of difference in CSP's. In *Proceedings AAAI'94*, pp. 362–367.

Rossi, F., van Beek, P., & Walsh, T. (2007). *Handbook of Constraint Programming*. Elsevier.

Sabin, D., & Freuder, E. (1994). Contradicting conventional wisdom in constraint satisfaction. In *Proceedings of ECAI-94*, pp. 125–129.

Schulte, C., & Carlsson, M. (2006). Finite domain constraint programming systems. In *Handbook of constraint programming*. Elsevier. Chapter 14.

Semple, C., Daniel, P., Hordijk, W., Page, R., & Steel, M. (2004). Supertree algorithms for ancestral divergence dates and nested taxa. *Bioinformatics*, *20*(15), 2355–2360.

Semple, C., & Steel, M. (2000). A supertree method for rooted trees. *Discrete Appl. Math.*, *105*(1-3), 147–158.

Smith, B. M. (1995). A Tutorial on Constraint Programming. Technical Report 95.14, University of Leeds.

TreeBASE (2003). http://www.treebase.org/ TreeBASE: a database of phylogenetic knowledge..

Tsang, E. (1993). *Foundations of Constraint Satisfaction*. Academic Press.

van Hentenryck, P., Deville, Y., & Teng, C.-M. (1992). A generic arc-consistency algorithm and its specializations. *Artificial Intelligence*, *57*, 291–321.

van Hentenryck, P., Saraswat, V., & Deville, Y. (1998). Design, implementation, and evaluation of the constraint language cc(fd). *Journal of Logic Programming*, *37*, 139–164.

Wu, G., You, J.-H., & Lin, G. (2007). Quartet-based phylogeny reconstruction with answer set programming. *IEEE/ACM Transactions on Computational Biology and Bioinformatics*, *4*, 139–152.

Yuanlin, Z., & Yap, R. H. C. (2001). Making AC-3 an optimal algorithm. In *IJCAI*, pp. 316–321.